\documentclass[lettersize,journal]{IEEEtran}
\usepackage{geometry}
\usepackage{fancyhdr}
\usepackage[export]{adjustbox}
\usepackage[utf8]{inputenc}
\usepackage{enumitem} 
\usepackage{amsmath}
\usepackage{tabulary}
\usepackage{enumitem}
\usepackage{lipsum}
\usepackage{tabularx}
\usepackage{adjustbox}
\usepackage{url}
\usepackage{hyperref}
\usepackage{amsmath,amssymb,amsfonts}
\usepackage{algorithm}
\usepackage{algpseudocode}
\usepackage{graphicx}
\usepackage{textcomp}
\usepackage{xcolor}
\usepackage{multirow}
\usepackage{tikz}
\usepackage{booktabs}
\usepackage{soul}

\usetikzlibrary{shapes, arrows, positioning, calc, backgrounds}

\makeatletter
\def\ps@customIEEEtitle{%
  \def\@oddhead{\footnotesize\hfil%
    \begin{tabular}{c}
      This is the author's version of the article published in IEEE Transactions on Power Delivery, Volume 40, Issue 1, February 2025. \\[1mm]
      The fully edited and published version is available on IEEE at: \url{http://doi.org/10.1109/TPWRD.2024.3486010}
      (Date of Publication: 24 October 2024)
    \end{tabular}%
    \hfil}%
  \def\@evenhead{\footnotesize\hfil%
    \begin{tabular}{c}
      This is the author's version of the article published in IEEE Transactions on Power Delivery, Volume 40, Issue 1, February 2025. \\[1mm]
      Fully edited version available at: \url{http://doi.org/10.1109/TPWRD.2024.3486010} \\[1mm]
      (Published online: 24 October 2024)
    \end{tabular}%
    \hfil}%
  \def\@oddfoot{\footnotesize\hfil%
    \begin{tabular}{l}
      © 2024 IEEE. Personal use of this material is permitted.
      Permission from IEEE must be obtained for all other uses, in any current or future media,\\[-0.4ex]
      including reprinting/republishing this material for advertising or promotional purposes, creating new collective works, for resale or redistribution to\\[-0.4ex]
      servers or lists, or reuse of any copyrighted component of this work in other works.
    \end{tabular}%
    \hfil}%
  \def\@evenfoot{\footnotesize\hfil%
    \begin{tabular}{c}
      © 2024 IEEE. Personal use of this material is permitted.\\[-0.8ex]
      Permission from IEEE must be obtained for all other uses, in any current or future media,\\[-0.8ex]
      including reprinting/republishing this material for advertising or promotional purposes, creating new collective works, for resale or redistribution to\\[-0.8ex]
      servers or lists, or reuse of any copyrighted component of this work in other works.
    \end{tabular}%
    \hfil}%
}
\makeatother

\begin{document}

\title{{Leveraging Hypernetworks and Learnable Kernels for Consumer Energy Forecasting Across Diverse Consumer Types}}

\author{Muhammad Umair Danish,~\IEEEmembership{Graduate Student Member,~IEEE}
        and Katarina Grolinger,~\IEEEmembership{Member,~IEEE \vspace{-15pt}}

\thanks{Muhammad Umair Danish and Katarina Grolinger are with the Department of Electrical and Computer Engineering, The University of Western Ontario, London, ON, Canada. E-mail: \{mdanish, kgroling\}@uwo.ca}
\thanks{Corresponding author: Katarina Grolinger (e-mail: kgroling@uwo.ca).}
\thanks{This work was supported by the Climate Action and Awareness Fund [EDF-CA-2021i018, Environnement Canada].}
\thanks{Manuscript received March 12, 2024; revised June 19, 2024, and September 27, 2024; accepted October 20, 2024}}
\markboth{IEEE TRANSACTIONS ON POWER DELIVERY,~Vol.~40, No.~1, February 2025}
{Danish \MakeLowercase{\textit{et al.}}: Leveraging Hypernetworks and Learnable Kernels for Consumer Energy Forecasting Across Diverse Consumer Types}

\maketitle
\thispagestyle{customIEEEtitle}

\begin{abstract}
Consumer energy forecasting is essential for managing energy consumption and planning, directly influencing operational efficiency, cost reduction, personalized energy management, and sustainability efforts. In recent years, deep learning techniques, especially LSTMs and transformers, have been greatly successful in the field of energy consumption forecasting. Nevertheless, these techniques have difficulties in capturing complex and sudden variations, and, moreover, they are commonly examined only on a specific type of consumer (e.g., only offices, only schools). Consequently, this paper proposes HyperEnergy, a consumer energy forecasting strategy that leverages hypernetworks for improved modeling of complex patterns applicable across a diversity of consumers. Hypernetwork is responsible for predicting the parameters of the primary prediction network, in our case LSTM. A learnable adaptable kernel, comprised of polynomial and radial basis function kernels, is incorporated to enhance performance. The proposed HyperEnergy was evaluated on diverse consumers including, student residences, detached homes, a home with electric vehicle charging, and a townhouse. Across all consumer types, HyperEnergy consistently outperformed 10 other techniques, including state-of-the-art models such as LSTM, AttentionLSTM, and transformer.
\end{abstract}


\vspace{-8pt}
\section{Introduction}
According to the United States Energy Information Administration, by the year 2050, the global energy demand is projected to increase by 50\% from its 2018 levels \cite{eia2023}. 
In the meantime, emphasis on sustainability and carbon footprint reduction has made energy management a focal point. Nations are setting climate action plans to counter the adverse effects of climate change and global warming: for example, the European Green Deal aims to achieve a 55\% reduction in carbon footprint and 32.50\% increase in energy efficiency by 2030 \cite{eu2022}. Energy forecasting can contribute to achieving these goals while providing savings: a mere 1\% improvement in forecast accuracy can result in an annual savings of \$1.6 million for energy-generating companies \cite{doe2012}. Consumer energy forecasting refers to predicting the future energy consumption patterns of individual consumers based on their historical usage data and other influencing factors such as weather. This type of forecasting differs from traditional load forecasting, which focuses on the overall demand across the grid, as it specifically targets consumer-level data to enhance personalized energy management and improve forecasting accuracy for diverse consumer types. Consumer energy forecasting has a wide range of applications including resource optimization, infrastructure planning, enhancing service quality, personalized energy solutions, and economic efficiency for both energy providers and consumers.

Despite advancements in consumer energy forecasting, accurately predicting energy demand in urban environments remains a challenge due to factors such as consumer behavior, changes in equipment, and variations in meteorological conditions \cite{chen2021application}. Machine Learning (ML) approaches, including support vector machines, artificial neural networks, and ensemble methods, have been extensively studied and consistently deliver good results in scenarios where strong data patterns are present \cite{aslam2021survey}. In recent years Deep Learning (DL) models, specifically Recurrent Neural Networks (RNNs), Long Short Term Memory Network (LSTM), Gated Recurrent Unit (GRU), and Transformers, have shown remarkable results due to their ability to capture long-range temporal dependencies \cite{lheureux2022transformer}.

However, the accuracy of these models can be compromised by difficulties in capturing complex energy consumption patterns and the variable nature of consumer behaviors, which are influenced by sudden events, new technology adoption, lifestyle changes, and meteorological shifts. These factors not only result in significant seasonality and cyclic variations in data but also introduce concept drift, novel dependencies, and elements of randomness. These changes can be challenging to predict \cite{cai2019day}.

To model energy patterns, in the training process, neural networks adjust their weights, most commonly through backpropagation or backpropagation through time. However, while these techniques optimize weights to minimize error, in the presence of complex and changing patterns, this optimization becomes challenging. Transfer learning also helps in learning and fine-tuning the weights but can be constrained by issues such as the domain gap and negative transfer \cite{das2023long, alqushaibi2020review}. Moreover, consumer energy forecasting studies mainly focused on a specific type of consumers such as residences \cite{gong2021peak}, apartments \cite{rezaei2020optimal}, detached houses \cite{kong2017short}, townhouses \cite{gong2021peak} and houses with electric vehicles \cite{zhang2020deep}. Nevertheless, there is a lack of studies aimed at developing forecasting strategies capable of handling a variety of consumers{, what is needed for practical applications.

In response to these challenges, this paper proposes HyperEnergy, an approach for {consumer energy} forecasting capable of handling complex patterns as well as accommodating diverse consumer types, thereby delivering a dependable forecasting strategy {that can be applied to diverse consumers to deliver consumer-specific models}. To optimize the weights, {HyperEnergy} takes advantage of hypernetworks, which are meta-networks designed to predict weights and biases for the primary network. The hypernetwork is altered by incorporating learnable adaptive kernels, consisting of polynomial and Radial Basis Function (RBF) kernels, to form a kernelized hypernetwork. The kernels are responsible for transforming data into higher dimensions while a learnable parameter controls the contribution of each kernel, assisting in modeling diverse and complex patterns. To accommodate the structure of a complex primary network, such as Long-Short-Term-Memory (LSTM), and its gating mechanism, the parameter integration module establishes connections between the hypernetwork and the primary network. 
The key contributions of this paper are as follows:  

\begin{enumerate}
    \item Design of {HyperEnergy} that incorporates LSTM as the primary network and the kernelized hypernetwork for learning the primary network weights. The two networks, connected through the parameter integration module, learn simultaneously with the objective of minimizing the prediction error. 
  
    \item Design and integration of modified polynomial and RBF kernels with the hypernetwork. A learnable parameter controls the contribution of each kernel, enabling the model to accommodate diverse patterns.   

    \item Examination of {HyperEnergy} demonstrating that it outperforms other techniques across diverse consumer types, including student residences, detached houses, townhouses, and houses with electric vehicles.
    
\end{enumerate}
The remainder of the paper is organized as follows: Section \ref{sec:relwork} presents the related work, Section \ref{sec:hyperforcasting} details the proposed {HyperEnergy}, Section \ref{sec:experiments} describes evaluation, and Section \ref{sec:results} presents results and analysis. Finally, Section \ref{sec:conclusion} concludes the paper.

\section{Related Work} \label{sec:relwork}
This section provides an overview of the significant advancements in {consumer energy} forecasting over the past five years followed by a discussion on hypernetworks. 

\subsection{ {Consumer Energy} Forecasting}
In the pursuit of improved {consumer energy} forecasting accuracy, numerous ML and DL approaches have been proposed. Support Vector Machines (SVMs) were incorporated into various solutions; however, their effectiveness greatly depends on the selection of kernels, which are responsible for handling non-linear data \cite{zhang2018forecasting}. The polynomial kernel is good at tracking gradual shifts but may falter with sudden spikes due to its constant polynomial degree \cite{madhukumar2022regression}. On the other hand, RBF kernel is a promising option for identifying sudden shifts in energy consumption; nevertheless, for success, it necessitates parameter tuning \cite{madhukumar2022regression}. Gradient boosting approaches, such as Extreme Gradient Boosting (XGBoost) and Light Gradient Boosting Machine (LightGBM), have also been proposed for {consumer energy} forecasting {\cite{ni2024light}}. However, these approaches encounter challenges in adapting to novel patterns, exhibit sensitivity to outliers, and may face scalability issues.

In recent years, DL models such as Multi-Layer Perceptrons (MLPs), Convolutional Neural Networks (CNNs), Recurrent Neural Networks (RNNs), Gated Recurrent Units (GRUs), LSTMs, and transformers, have extensively been studied for {consumer energy} forecasting. MLPs may fall short in recognizing temporal relationships within the energy data. To address this issue, advanced architecture such as the Time-series Dense Encoder (TiDE) \cite{das2023long} and Neural Basis Expansion Analysis (N-BEATS) \cite{oreshkin2021n} have been suggested. TiDE combines the simplicity of linear models with a temporal encoder which makes it a promising approach for long-term forecasting \cite{das2023long}. N-BEATS is designed with a series of MLP stacks and blocks to provide interpretable time-series forecasting which offers a fresh perspective for short-term forecasting \cite{oreshkin2021n}. Its complex architecture requires a significant amount of data for training. While CNNs excel in processing spatial patterns, they encounter challenges in handling temporal dynamics \cite{shaikh2023new}.

RNNs, GRUs, and LSTMs are known for their ability to handle sequential and time-series data. Specifically, LSTMs are seen as a highly suitable choice for {consumer energy} forecasting \cite{yamak2019comparison} due to the gating structures that enable LSTM to capture long-term dependencies. This ability is important in short-term {consumer energy} forecasting, where LSTM can capture sudden changes \cite{kong2017short}. Skala et al. proposed LSTM Bayesian neural networks for interval {consumer energy} forecasting for individual households in the presence of electric vehicle charging \cite{skala2023interval}.

Transformers and their advanced counterparts have contributed immensely to prediction capabilities in the energy sector \cite{vaswani2017attention}. They bring innovative self-attention mechanisms to handling complex data patterns. L’Heureux et al. proposed a transformer-based architecture and examined it on an open-source dataset comprising of 20 zones from a US utility company. The results showed that the transformer outperformed LSTM and sequence-to-sequence model \cite{lheureux2022transformer}. Moreover, the emergence of hybrid models that combine machine learning, statistical, and deep learning models has been notable in {consumer energy} forecasting \cite{li2023short, li2023novel, triebe2019ar}.

While ML and DL models have been greatly successful in {consumer energy} forecasting, they encounter challenges in learning optimal weights when dealing with sudden spikes, drops, concept drift, or level shifts. This may lead to reduced forecasting accuracy. Hypernetworks have the potential to remedy this by assisting the primary network to learn weights. 

Moreover, the existing literature predominantly concentrates on particular consumer types, neglecting to offer a generic solution suitable for diverse consumer groups.  For instance, Lin et al. \cite{lin2022hybrid} used residential data, Rezaei et al. \cite{rezaei2020optimal} concentrated on apartments, Kong et al. \cite{kong2017short} worked with individual houses, Gong et al. \cite{gong2021peak} focused on townhouses, and Zhang et al. \cite{zhang2020deep} explored energy patterns in houses equipped with electric vehicles. This gap leads to a question: Can a single forecasting model successfully adapt to and capture the diverse energy consumption patterns observed across various consumer groups? Therefore, our study proposes a solution based on hypernetworks to facilitate modeling complex patterns and demonstrates that the proposed {HyperEnergy} achieves superior performance compared to other techniques across a variety of consumers.

\subsection{Hypernetworks}

Hypernetworks are meta neural networks that generate weights and biases for another neural network known as the primary network. In this arrangement, the hypernetwork's outputs, weights and biases for the primary network, are received by the primary network \cite{hoopes2021hypermorph} and the primary network then utilizes these parameters to execute its tasks. The two networks are trained simultaneously and the hypernetwork customizes the primary network's parameters based on its inputs. Initially, hypernetworks were designed to compress neural network sizes \cite{schmidhuber1993self} but have now found many applications including network pruning \cite{li2020dhp}, multitask learning \cite{meyerson2019modular}, functional representation \cite{klocek2019hypernetwork}, and generative tasks \cite{ratzlaff2019hypergan}.

In HyperMorph \cite{hoopes2021hypermorph}, three distinct hypernetwork-based learning strategies for image registration were investigated: pre-integrative, where input is provided at the beginning of the primary model; post-integrative, involving input into the final layers; and fully-integrative, which monitors the entire model. The pre-integrative learning strategy yields better results than the remaining two techniques \cite{hoopes2021hypermorph}.

Hypernetworks have been explored and achieved notable success in various domains. For example, in recommendation systems, hypernetworks have been integrated to address the user cold-start problem \cite{lu2023hyperrs}. In classification tasks, hypernetworks have been merged with graph networks and transformers to improve the classification of graph structures. Additionally, hypernetworks have proven highly effective in addressing differential privacy issues within the field of federated learning \cite{nemala2023differential}.

Despite their successes in different fields, hypernetworks' potential remains largely unexplored in {consumer energy} forecasting. Primarily, hypernetworks have been used with feedforward neural networks and CNNs \cite{hoopes2021hypermorph} which are not well suited for {consumer energy} forecasting as they are not specifically designed for capturing temporal dependencies. Consequently, our study proposes a hypernetwork with learnable adaptive kernels and LSTMs for {consumer energy} forecasting, aimed at handling time dependencies and accommodating the diverse consumer groups.

\vspace{-5pt}
\section{{HyperEnergy}} \label{sec:hyperforcasting}

This section presents {HyperEnergy}, our proposed method for {consumer energy} forecasting, designed to accommodate a wide range of energy consumers. The overview is depicted in Figure \ref{fig:deep_learning_arch}, while the three main components, kernelized hypernetwork, parameter integration module, and the primary network, are described in the following subsections, followed by a discussion of the parameter update process.

\begin{figure*}[h!]
    \centering
    \includegraphics[width=0.85\textwidth]{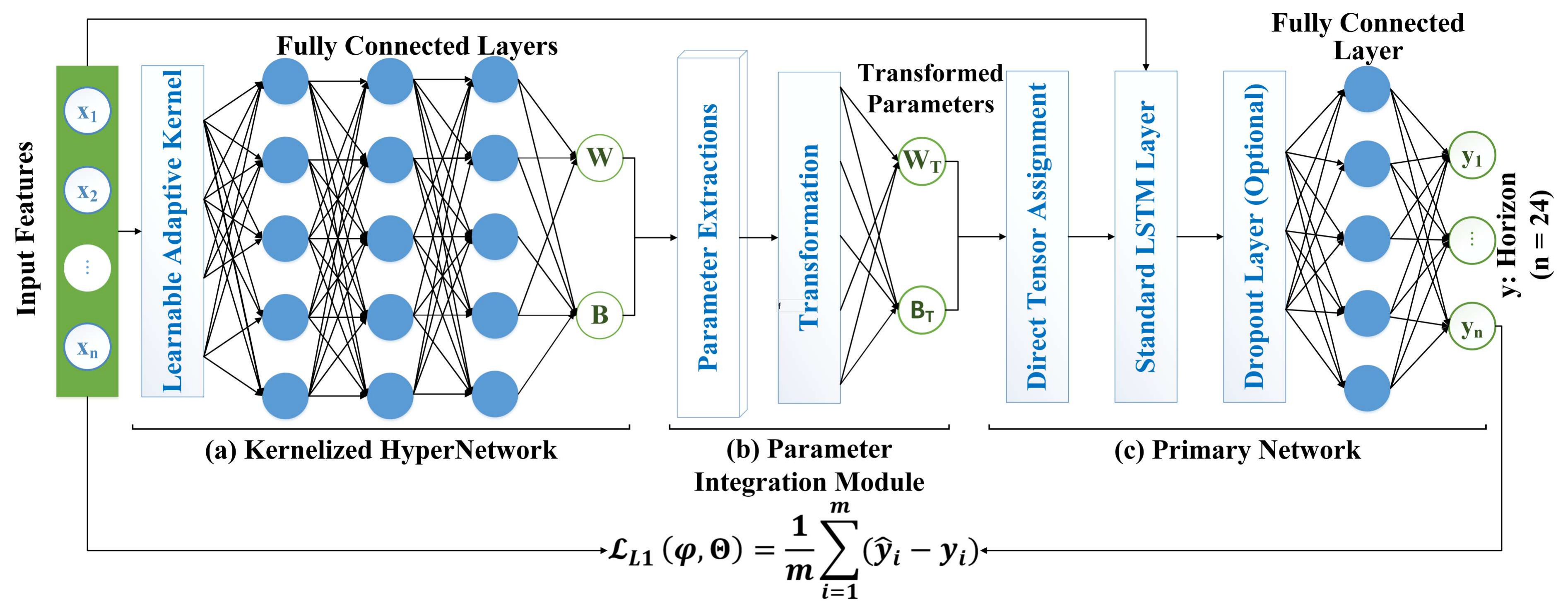}
    \vspace{-8pt}
    \caption{The proposed {HyperEnergy}, a deep learning technique, consists of three main components: (a) the kernelized hypernetwork, which contains learnable adaptive kernels, fully connected layers, and predicts weights and biases; (b) the parametric integration module, responsible for extracting and transforming weights and biases to ensure compatibility with LSTM; and (c) the primary network, consisting of LSTM and fully connected layers responsible for generating the final outputs.}  
    \label{fig:deep_learning_arch}
    \vspace{-5pt}
\end{figure*}
\vspace{-5pt}
\subsection{Kernelized Hypernetwork}

The kernelized hypernetwork denoted as \(H_k\) is designed to generate the weights and biases represented as \( \Theta \) for the primary network. The input to this \(H_k\) network is \(x \in \mathbb{R}^{m \cdot k \cdot n}\) where \(m\) is the number of samples, \(k\) is the number of features, and \(n\) is the number of time-steps in a sample as generated through the sliding window technique \cite{lheureux2022transformer}. Here, features \(k\) encompass attributes such as temperature, day of the year, day of the month, day of the week, hour of the day, and energy consumption from the preceding \( n \) hours. This input is passed to the learnable adaptive kernel followed by the fully connected layers. 
{The Kernelized HyperNetwork generates as output $\Theta$, which represents the parameters of the primary network.}

\subsubsection{Learnable Adaptive Kernel}
{The role of the learnable adaptive kernel is to transform input features into a high-dimensional space, thereby helping to capture both gradual and sudden changes in {energy consumption} values. The learnable adaptive kernel provides these transformed features as output.} Kernels are well-known for their ability to transform data into high-dimensional spaces, which helps the model capture variations in patterns \cite{kouziokas2020svm}. A learnable adaptive kernel is the first part of the kernelized hypernetwork; it is a learnable combination of polynomial and RBF kernels responsible for transforming data before the fully connected layers {\cite{zhang2018forecasting}}. 
{In contrast to traditional kernel-based forecasting methods that take input features and directly predict output values \cite{zhang2018forecasting, madhukumar2022regression}, {HyperEnergy} incorporates kernels into a hypernetwork to predict the parameters of the primary network. Moreover, we designed learnable kernels and combined two types of kernels, which, together with the hypernetwork, provide {HyperEnergy} with flexibility and applicability across diverse consumers.}

A traditional polynomial kernel is defined as follows:
\begin{equation}
%K_{\text{poly}}(x, x') = \left(c + x^T x' \right)^d 
K_{\text{poly}}(x, x') = \left(x^T x' + c \right )^d 
\end{equation}
\noindent Here, \( x \) and \(x'\) represent two data points, \( c \) is a constant term, and \( d \) denotes the degree of the polynomial. In contrast to traditional kernel approaches which compute the distance between all pairs of samples \(x\)  and \(x'\), we employ learnable reference points \( r_j \) in place of \(x'\) for enhancing models adaptability to diverse data patterns. The reference points \( r_j \), where \( j = 1, 2, \ldots, N_r \), with \( N_r \) representing the total number of reference points, are initialized randomly. This initialization is formulated as:
\begin{equation}
r_j = \rho(N_r, k \cdot n)
\end{equation}
\noindent Here, \( \rho \) generates values from a normal distribution, \( N_r \) is the number of reference points, \( k \) denotes the number of features, and \( n \) is the number of time steps. Each reference point \( r_j \) is thus a vector in \( \mathbb{R}^{k\cdot n} \), initialized with random values. As the training progresses, these reference points are updated based on the gradient. The updating process at each training step can be represented as:
\begin{equation}
r_j^{(new)} = r_j^{(old)} - \eta \cdot \frac{\partial \mathcal{L}}{\partial r_j},
\end{equation}
\noindent where \( r_j^{(old)} \) and \( r_j^{(new)} \) represent the reference points before and after the update, respectively, \( \eta \) is the learning rate, and \( \frac{\partial \mathcal{L}}{\partial r_j} \) is the gradient of the loss function \( \mathcal{L} \) with respect to the reference point \( r_j \). 
{The loss function, as elaborated upon in Subsection \ref{sec:bk}, can be either Mean Absolute Error (MAE) or Mean Squared Error (MSE), selected through hyperparameter optimization.} 

Building upon the described reference points, {the improved kernel takes $x$ and the reference points $r_j$ learned during training as inputs to produce transformed features.  
We} define a learnable version of the polynomial kernel as follows:
\begin{equation}
K_{\text{p}}(x, r_j) = \left(\alpha \cdot x r_j^T + c \right)^d
\label{eq:p}
\end{equation}
Here, \( \alpha \) represents the scale factor, \( c \) is the constant term, and \( d \) denotes the degree of the polynomial. 

Similarly, {an improved form of the RBF kernel is defined for the same set of input points $x$ and learned reference points $r_j$ to create transformed features}, with the parameter \( \gamma \) controlling the spread of the kernel:
\begin{equation}
K_{\text{r}}(x, r_j) = \exp\left(-\gamma \cdot \|x - r_j\|^2 \right)
\label{eq:r}
\end{equation}

\noindent {As with the polynomial kernel, reference points are learned through the training process.}

Finally, we define the learnable adaptive kernel (LAK) by integrating kernels described in equations \ref{eq:p} and \ref{eq:r}, and introducing a parameter \(\lambda\) to control the contribution of each kernel during training.
%Here, Learnable Adaptive Kernel is defined as, 
\begin{equation}
K_{\text{o}} = \lambda K_{\text{p}}(x, r_j) + (1 - \lambda) K_{\text{r}}(x, r_j)
\end{equation}
The parameter \(\lambda\), constrained within the range [0, 1], regulates the contribution of each kernel and is learned during the training process. The RBF kernel is suited for identifying sudden spikes or drops in energy data due to its local sensitivity and the adaptability of the \(\gamma\) parameter, while the polynomial kernel is good at capturing gradual changes {\cite{madhukumar2022regression}},{\cite{zhang2018forecasting}}. Combining the strengths of both, the learnable adaptive kernel assists the model in capturing complex data patterns by introducing another learning layer. This layer allows the hypernetwork to fine-tune its response, thereby providing better parameters to the primary network and improving the predictions. 

\subsubsection{Fully Connected Layers} 
The transformed features from the learnable adaptive kernel are processed through fully connected layers, with the activation function applied on each layer. This operation is represented as:
\begin{equation}
\textit{l}_1(x) = W_1 \cdot K_{\text{o}} + b_1
\end{equation}
\begin{equation}
\textit{a}_1(x) = a(\textit{l}_1(x))
\end{equation}
\begin{equation}
\textit{l}_2(x) = W_2 \cdot \text{a}_1(x) + b_2
\end{equation}
\begin{equation}
\textit{a}_2(x) = a(\textit{l}_2(x))
\end{equation}
\noindent where \(W\) and \(b\) are the weights and biases in the network. The activation function, ReLU or Swish \cite{ramachandran2017swish}, is selected through hyperparameter optimization. Similarly, the number of fully connected layers in the hypernetwork is also determined through hyperparameter optimization. The final layer of the kernelized hypernetwork responsible for generating the parameters \( \Theta \) required by the primary network, is given by:
    \begin{equation}
        \Theta = W_3 \cdot \text{a}_2(x) + b_3
    \end{equation} 

The parameters \( \Theta \) are passed to the parameter integration module for merging with the primary network.
\vspace{-0.3 cm}
\subsection{Parameter Integration Module}
{The purpose of the parameter integration module is to transform the parameters \( \Theta \) to be compatible with the LSTM internal gating structure and the assignment of weights and biases.} The parameters provided by the hypernetwork cannot be directly pushed to LSTM due to its gating mechanism. To address this, we apply a transformation on \(\Theta\). The output of the kernelized hypernetwork \(\Theta\) is a tensor with a shape equal to one $\times$ the number of parameters in the primary network. 

We exclusively extract and transform the parameters needed for the LSTM layer as the objective is to optimize solely the weights and biases of the LSTM layer, which is the primary forecasting component. The parameters of the fully connected layer are not required in this context, as this layer functions merely for output purposes and relies on the representations learned in the LSTM layer.
The weights and biases for the LSTM are obtained through tensor slicing and transformed from \(\Theta\) as follows:
    \begin{equation}
        W_{\text{T}} = \mathcal{T}(\Theta, p_{wi}, p_{we}, [4u, k \cdot v + u])
    \label{eq:w}
    \end{equation}
\noindent where \( p_{wi} \) and \( p_{we} \) represent the starting and ending indices for LSTM weights within the entire parameter tensor. In our experiments \( p_{wi} \) is set to $0$, thus, \( p_{we} \) equals the number of parameters in LSTM. This indicates the extraction of only the LSTM parameters from the entire set of parameters.

\(\mathcal{T}\) transforms the extracted weights to match the LSTM's weight shape \([4u, k \cdot v + u]\). Here, \(4u\) represents the units across the four gates of the LSTM —- input, forget, cell, and output, each with \(u\) units. The term \(k \cdot v\) indicates the input dimension, where \(k\) is the number of features and \(v\) is the number of time steps. 
Similarly, biases are extracted as: 
    \begin{equation}
        B_{\text{T}} = \mathcal{T}(\Theta, p_{bi}, p_{be}, [4u])
    \label{eq:b}
    \end{equation}
where \( p_{bi} \) and \( p_{be} \) are the start and end indices for LSTM biases. The dimension \([4u]\) represents the total size of the biases for the LSTM. 

\vspace{-0.3 cm}
\subsection{Primary Network}
{The role of the primary network is to receive the parameters $\Theta$ from the parameter integration module and input features, and generate future {energy consumption} values as output.}
We selected LSTM as the primary network due to its success in modeling temporal data. Specifically, the primary network consists of a standard LSTM layer(s) and a fully connected output layer. The LSTM's internal parameters, including weights and biases, are directly updated with the tensors \( W_{\text{T}} \) from Equation (\ref{eq:w}) and \( B_{\text{T}} \) from Equation (\ref{eq:b})   which are the outputs of the transformation process from the parameters integration module. 

Direct tensor assignment, denoted as \(\Psi\), updates weights and biases by directly assigning the relevant portions from the hypernetwork's outputs. This operation is performed in a way that guarantees that the LSTM parameter update does not interfere with the ongoing training gradients and backpropagation process. In other words, this operation assigns LSTM parameters in a gradient-free manner ensuring that LSTM layers are excluded from gradient updates in the backpropagation process:
\begin{equation}
\Psi(W_{\text{T}}, B_{\text{T}}) \rightarrow \text{LSTM}
\end{equation}
In addition to setting the parameters using \(W_\text{T}\) and  \(B_\text{T}\) from \(H_k\), the primary network also takes the same inputs \(x \in \mathbb{R}^{m \cdot k \cdot n}\) as the hypernetwork and produces \(h_t\), \(c_t\):
\begin{equation}
    h_t, c_t = \text{LSTM}(x, W_{\text{T}}, B_{\text{T}})
\end{equation}
where \( h_t \) and \( c_t \) are the hidden state and the cell state of the LSTM at time step \( t \). 

The hidden state \(h_t\) is passed to the fully connected layer to generate the final {energy consumption} predictions for the next \(h \) time steps. The fully connected layer is represented as:
\begin{equation}
    \hat{y}_h = W_{\textit{l3}}\cdot h_t + b_{\textit{l3}},
\end{equation}
\noindent where \(W_{\textit{l3}}\) and \(b_{\textit{l3}}\) are the weights and biases and \(\hat{y}_h\) is the vector of predicted energy consumption for the next \(h\) hours.

\subsection{Backpropagation and Parameter Update Process}
\label{sec:bk}
The MSE or MAE loss functions, based on the results of hyperparameter optimization, are used as the loss functions to quantify the difference between the predictions and the actual target values. They are calculated as:
\begin{equation}
    \mathcal{L}_{MSE}( \Phi, \Theta) = \frac{1}{m} \sum_{i=1}^m (\hat{y}_i - y_i)^2,
    \label{sec:l1}
\end{equation}
\begin{equation}
    \mathcal{L}_{MAE}( \Phi, \Theta) = \frac{1}{m} \sum_{i=1}^m |\hat{y}_i - y_i|,
    \label{sec:l2}
\end{equation}
%where, \( \hat{y}_i \) is the predicted output from the Primary Network, and \( y_i \) is the true output.
\noindent where, \( \hat{y}_h \) is the predicted output from the primary network, \( y_h \) is the actual {energy consumption} value, and \(\Phi\) are kernelized hypernetwork parameters. 

It is important to note that MAE is more robust to outliers than MSE because MSE squares the errors, amplifying the impact of outliers. However, MSE provides a smooth gradient for optimization, allowing for controlled updates during training. Therefore, the selection between the two is carried out as part of the hyperparameter optimization process. 

After the loss is calculated, the backpropagation process differs from traditional backpropagation. Traditional backpropagation calculates gradients with respect to the weights and biases of the prediction network, and then those weights and biases are updated. In contrast, in our approach, gradients are calculated with respect to the hypernetwork parameters, and only the hypernetwork parameters are updated through backpropagation. The primary prediction network, LSTM, is not updated through backpropagation; instead, it receives its parameters from the hypernetwork during the forward pass. Although deep learning is employed, our model uniquely uses a hypernetwork with learnable kernels to predict parameters for the primary network, unlike traditional neural networks that learn weights and biases directly through backpropagation.

\section{Evaluation} \label{sec:experiments}
This section first describes datasets, preprocessing, and performance metrics. Next, hyperparameter optimization and the architectures included in the comparison are discussed.

\subsection{Datasets, Preprocessing, and Evaluation Metrics}
The evaluation utilized ten distinct real-world datasets from two primary consumer groups: student residences and individual homes. The overview of the datasets presented in Table \ref{tab:dataset_description} includes the time frames during which data were collected and a brief description of each dataset. Within each category, there are major differences between the buildings. Residence 1 offers suite-style accommodation with a shared kitchen, while Residence 2 adopts suite style but without a kitchen. Both residences accommodate over 400 students. 

While all considered homes are located in London, Ontario, Canada, there is significant diversity among them. Homes 1, 2, and 3 are all detached properties, but Home 3 stands out due to the presence of an electric vehicle, leading to notable {energy consumption} fluctuations caused by at-home charging. Home 4 is a 3-bedroom townhouse, and thus, energy consumption will differ from detached homes because of the impact of neighboring units. {To investigate a diversity of non-residential consumers, a manufacturing building, a medical clinic, a retail store, and an office building are also considered \cite{Miller2020-yc}.}

\begin{table}[!t]
\vspace{-10pt}
\centering
\caption{Description of Energy Consumption Datasets}
\label{tab:dataset_description}
\setlength{\tabcolsep}{3pt}
\begin{tabular} { l l p{3.5cm}  }
\toprule
\textbf{Dataset} & \textbf{Dates} & \textbf{Short Description} \\
\midrule
\multicolumn{2}{l} {\textbf{Student Residences}}\\
\hspace*{0.5em} Residence 1 & Jan/2019 - Jul/2023 & A suite-style residence with shared kitchen \\ 
\hspace*{0.5em} Residence 2 & Jan/2019 - Jul/2023 & A suite-style residence without a kitchen \\
\midrule
\multicolumn{2}{l} {\textbf{Individual Houses}}\\

\hspace*{0.5em}House 1 & Jan/2002 - Dec/2004 & A detached home with complex energy usage patterns \\

\hspace*{0.5em}House 2 & Mar/2021 - Aug/2021 & A 2-bedroom detached house  \\

\hspace*{0.5em}House 3  & Mar/2021 - Aug/2021 & A 2-bedroom detached house with an electric vehicle \\

\hspace*{0.5em}House 4 & Mar/2021 - Aug/2021 & A 3-bedroom townhouse \\
\midrule

\multicolumn{2}{l} {{\textbf{Industrial and Commercial}}}\\

\hspace*{0.5em} {Manufacturing} & {Jan/2016 - Dec/2017} & {A manufacturing unit} \\

\hspace*{0.5em} {Medical Clinic} & {Jan/2016 - Dec/2017} & {A medical and wellness clinic} \\

\hspace*{0.5em} {Retail Store}  & {Jan/2016 - Dec/2017} & {A retail store} \\

\hspace*{0.5em} {Office} & {Jan/2016 - Dec/2017} & {A dedicated building for \mbox{offices}} \\
\bottomrule
\end{tabular}
\end{table}

Each dataset contained a recording date/time with corresponding hourly energy consumption. From the date/time, we extracted additional features including the day of the year, the day of the month, the day of the week, and the hour of the day to assist in modeling seasonal, weekly, and daily patterns. To capture weather patterns and enhance prediction accuracy, temperature data was incorporated and additional relevant features can be integrated if available. The data were normalized using Min-Max scaling to reduce the dominance of the large features and improve convergence.

Each dataset was divided into training, validation, and test sets with 60\%-20\%-20\% ratio. As data are temporal, it was prepared for the models using a sliding window technique with a widow length of 24 and a stride of 1. All models take as the input the previous 24 hours of five features including energy {energy consumption} and predict the next 24 hours consumption as output. This forecasting length was selected as energy operations commonly rely on the next day forecasts for energy planning. 

The evaluation was conducted using three metrics commonly employed in consumer energy forecasting: Mean Absolute Error (MAE), which measures the average absolute difference between predicted and actual values \cite{fekri2023asynchronous}; Root Mean Square Error (RMSE), which provides the standard deviation of the prediction errors (residuals) \cite{fekri2023asynchronous, l2022transformer}; and Symmetric Mean Absolute Percentage Error (SMAPE), which expresses the forecasting error as a percentage, facilitating easy interpretation and enables comparison across datasets \cite{fekri2023asynchronous}. SMAPE was selected over Mean Absolute Percentage Error (MAPE) as MAPE is biased toward large values and becomes undefined if there are actual values of zero. The SMAPE metric is calculated as:

{\begin{equation}
    \text{SMAPE} = 100\% \times \frac{1}{m} \sum_{i=1}^m \frac{2 |y_i - \hat{y}_i|}{|y_i| + |\hat{y}_i|},
\end{equation}
\noindent where \(y_i\) and \(\hat{y}_i\) are the actual and predicted {energy consumption} values, respectively, and \(m\) is the number of samples.}

\vspace{-2 mm}
\subsection{Hyperparameter Optimization}

To ensure fair treatment of all compared models, including ours,  hyperparameter optimization using the grid search was conducted for each dataset and each model. The hyperparameter search space is shown in Table \ref{tab:hyperparam_range_non_transformer}. Some of the hyperparameters are model-specific (e.g., attention head count), and this information is also included in the table.

\begin{table}[!b]
\vspace{-12 pt}
\caption{Hyperparameter Search Space for all Models}
\centering
\begin{tabular}{ll}
\toprule
\textbf{Hyperparameter}           & \textbf{Range of Values}    \\ \midrule 
Size of Hidden Layer (all models)         & 64, 128, 256            \\ 
Choice of Optimizer (all models)          & Adam, SGD, AdamW            \\ 
Attention Head Count (transformer only)         & 2, 4, 6                      \\ 
Objective Function (all models)           & MAE, MSE                    \\ 
Polynomial Degree (our model only)           & 2, 3, 4, 5                  \\ 
RBF Kernel Coefficient (our model only)       & 2,5,6,8,10                   \\ 
Activation Function (our model only)          & ReLU, Swish                   \\ 
\bottomrule
\end{tabular}
\label{tab:hyperparam_range_non_transformer}
\end{table}

\begin{table}[!b]
\vspace{-7 pt}
\centering
\caption{Selected Hyperparameters for {HyperEnergy}}
\begin{tabular}{p{1.8cm}p{1.30cm}p{0.7cm}p{0.7cm}p{1cm}p{0.65cm}}
\toprule 
\textbf{Dataset}  & \textbf{Hidden Units}   & \textbf{\(\gamma\)} & \textbf{\(d\)}  & \textbf{Optimizer} & \textbf{Loss} \\  \midrule 
Residence 1  & 256  & 1 & 2  & Adam & MAE  \\
Residence 2  & 128  & 1 & 3  & Adam & MAE  \\
House 1  & 128  & 10 & 2  & SGD & MAE  \\
House 2  & 128  & 5 & 2  & SGD & MSE  \\
House 3  & 128 & 6 & 2  & SGD & MAE \\
House 4 & 128  & 2 & 2  & SGD & MAE \\
\bottomrule
\end{tabular}
\label{tab:model_spec}
\vspace{-7 pt}
\end{table}

The early stopping mechanism monitored the validation loss, terminating training after five consecutive epochs without improvement. The maximum number of epochs was set to 300, with possible termination earlier if the early stopping condition was met. The learning rate was optimized using the ReduceLROnPlateau schedule \cite{torghabeh2023effectiveness}, which adjusts the rate during training if there is no performance improvement. The weights were initialized using uniform Xavier initialization to facilitate training \cite{kumar2017weight}.

Our implementation of the {HyperEnergy} included two hidden layers in LSTM and after hyperparameter optimization, for each of the ten models corresponding to ten datasets, the selected activation function was \(swish\). The remaining hyperparameters selected in the grid search differ among datasets, as depicted in Table \ref{tab:model_spec}.

\subsection{Forecasting Techniques Included in Comparison}

The proposed {HyperEnergy} was compared to 10 different {consumer energy forecasting} techniques, as listed in Table \ref{tab:comparison}. The table includes the base architecture together with pertinent literature references. To conduct a comprehensive analysis, the comparison includes diverse architectures encompassing a broad spectrum of predictive capabilities, starting from the basic Multi-Layer Perceptron (MLP) and advanced fully connected architecture, Neural Basis Expansion Analysis for interpretable Time Series forecasting (N-BEATS) and  Autoregressive Feed-forward Neural Network (ARFFNN). These techniques are followed by Temporal Convolution Neural Network (TempConv) and Extreme Gradient Boosting (XGBoost). Four variants of the recurrent networks are included: vanilla RNN (RNN), LSTM, GRU, and LSTM with attention mechanism (AttentionLSTM). Finally, the transformer was also examined.

\begin{table}[!t]
\caption{Forecasting Techniques Included in Comparison}
\centering
\begin{tabular}{p{0.5cm} p{2cm} p{5cm}}
\toprule
\textbf{No.} & \textbf{Model} & \textbf{Architecture} \\ \midrule 
1 & MLP & Three-layer fully connected network. \\ 
2 & N-BEATS & N-BEATS \cite{oreshkin2021n}. \\ 
3 & ARFFNN & AR-Net for Time-Series \cite{triebe2019ar}. \\ 
4 & Temporal Convolution & Two 1x1 convolutional layers with fully connected layer. \\ 
5 & XGBoost & Standard XGBoost. \\
6 & RNN & Vanilla RNN with fully connected layer. \\ 
7 & GRU & Standard GRU with fully connected layer. \\ 
8 & LSTM & Standard LSTM with fully connected layer. \\ 
9 & Attention LSTM & Attention-LSTM for load forecasting \cite{qin2022multi}. \\
10 & Transformer & Attention Is All You Need \cite{vaswani2017attention}. \\ 
 \bottomrule
\end{tabular}
\label{tab:comparison}
\vspace{-7 pt}
\end{table}

\section{Results and Analysis} \label{sec:results}
This section presents the results and discusses the findings: first for residences and then for the individual homes.

\subsection{Student Residences}

\label{sec:analysis_residences}
The two student residences have somewhat similar energy consumption patterns likely due to the similarity in students' daily routines and study habits. As seen from Figures \ref{fig:essex_energy} and \ref{fig:perth_energy}, there is a noticeable mix of trends  {and seasonality in energy usage across the training, validation, and test sets that correspond with the academic calendar and student occupancy.}
The changes occur at the start of the summer term when the majority of students are not on campus. However, as the considered period spans the COVID-19 restrictions, these patterns are less prominent and data have unexpected fluctuations. This imposes challenges in {consumer energy forecasting} as seen from the forecasting results.

\begin{figure}[!t]
    \centering
    \includegraphics[width=\columnwidth]{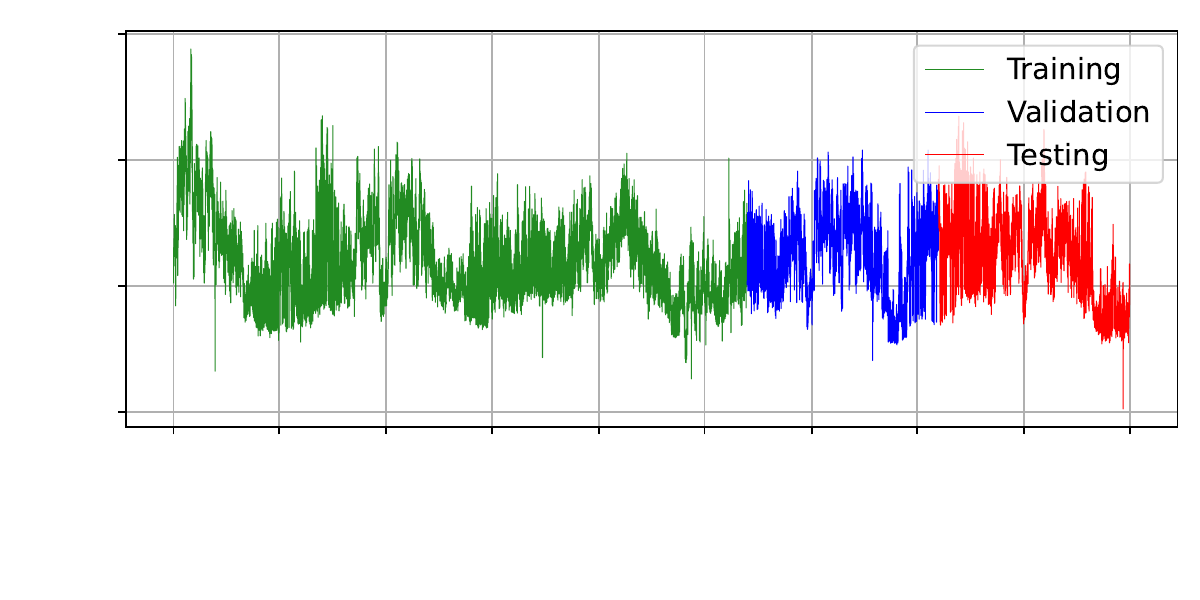} 
    \vspace{-25pt}
    \caption{Student Residence 1: {energy consumption} characterized by observable seasonal variations affected by students' routines.}
    \label{fig:essex_energy}

\vspace{5pt}
    \centering
    \includegraphics[width=\columnwidth]{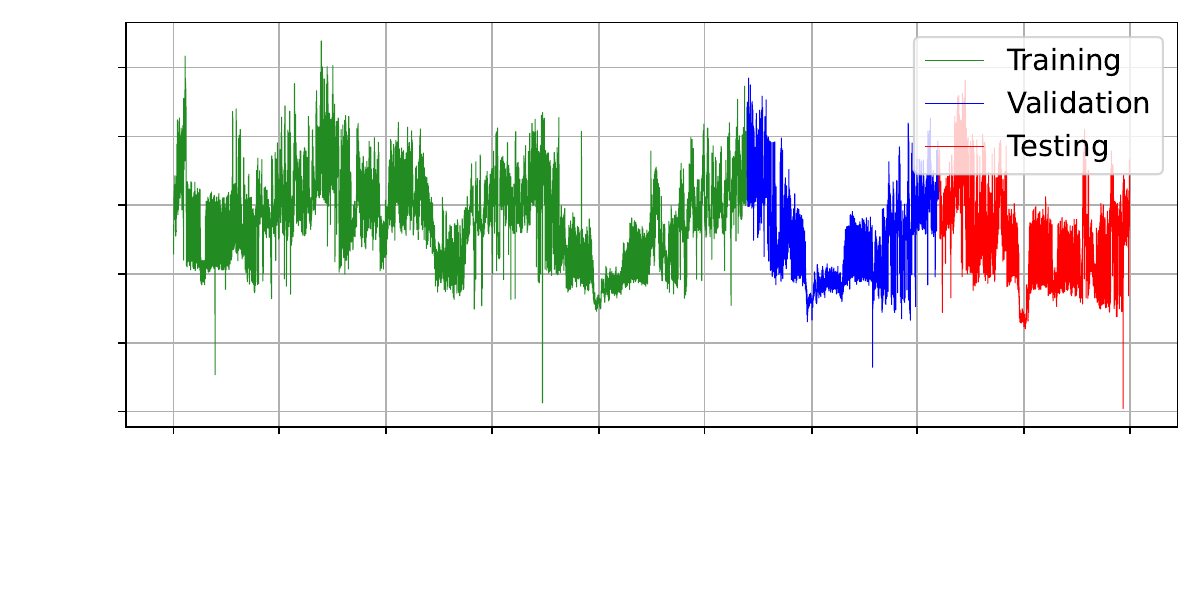}  
    \vspace{-25pt}
     \caption{Student Residence 2: {energy consumption} characterized by observable seasonal variations affected by students' routines.}
    \label{fig:perth_energy}
    \vspace{-7pt}
\end{figure}

\begin{table}[!t]
\centering

%\caption{Performance Comparison for the Student Residences}
\caption{Performance Comparison for the Student Residences {(Testing: 335 days)}}

\label{tab:performance_metrics_residences}
\begin{tabular}
{@{\hspace{1.2mm}}l@{\hspace{1.2mm}}|@{\hspace{1.2mm}}r@{\hspace{1.2mm}}r@{\hspace{1.2mm}}r|r@{\hspace{1.2mm}}r@{\hspace{1.2mm}}r}
\toprule
\multirow{2}{*}{\textbf{Model}} & \multicolumn{3}{c|}{\textbf{Residence 1}} & \multicolumn{3}{c}{\textbf{Residence 2}} \\
\cmidrule{2-7}
& \textbf{MAE} & \textbf{RMSE} & \textbf{SMAPE}  & \textbf{MAE} & \textbf{RMSE} & \textbf{SMAPE} \\
\midrule
MLP & 29.60 & 39.16 & 12.31\% & 29.24 & 36.89 & 11.61\% \\ 
NBEATS & 25.63 & 35.45 & 10.67\% & 20.60 & 29.87 & 8.43\% \\ 
ARFFNN & 24.08 & 32.58 & 9.97\% & 19.19 & 26.65 & 7.82\%\\ 
TempConv & 23.47 & 31.09 & 10.16\% & 33.20 & 44.63 & 13.56\%\\ 
XGBoost & 22.58 & 30.35 & 9.32\% & 18.68 & 25.90 & 7.60\% \\
RNN & 24.66 & 34.82 & 10.13\% & 29.05 & 39.50 & 12.00\% \\ 
GRU & 24.85 & 33.00 & 10.47\% & 19.67 & 28.04 & 8.14\%\\ 
LSTM & 21.82 & 29.85 & 9.12\% & 21.76 & 30.33 & 9.29\%\\ 
AttentionLSTM & 22.31 & 30.51 & 9.24\% & 18.23 & 27.33 & 7.51\% \\ 
Transformer & 22.46 & 31.09 & 9.41\%  & 29.06 & 36.59 & 11.52\% \\ 
 \textbf{{HyperEnergy}} & \textbf{20.02} & \textbf{27.58} & \textbf{8.27\%} & \textbf{16.49} & \textbf{24.59} & \textbf{6.70\%} \\ \bottomrule
\end{tabular}
%\vspace{-8pt}
\end{table}

The results of the comparison between our {HyperEnergy} and other state-of-the-art models are presented in Table \ref{tab:performance_metrics_residences}. {HyperEnergy} consistently outperforms traditional and advanced forecasting models in terms of all three metrics: MAE, RMSE, and SMAPE. For Residence 1, {HyperEnergy} achieves MAE of 20.02, RMSE of 27.58, and SMAPE of 8.27\%, while for Residence 2, all error metrics are lower: MAE is 16.49, RMSE is 24.59, and SMAPE is 6.70\%. Both, the standard LSTM and its hybrid variant, Attention LSTM, along with Transformer, ARFFNN, and XGBoost also demonstrate commendable performance by recording SMAPE values of less than 10\% for both residences. 

\begin{figure}[!t]
    \centering
    \vspace{-10pt}
    \includegraphics[width=\columnwidth]{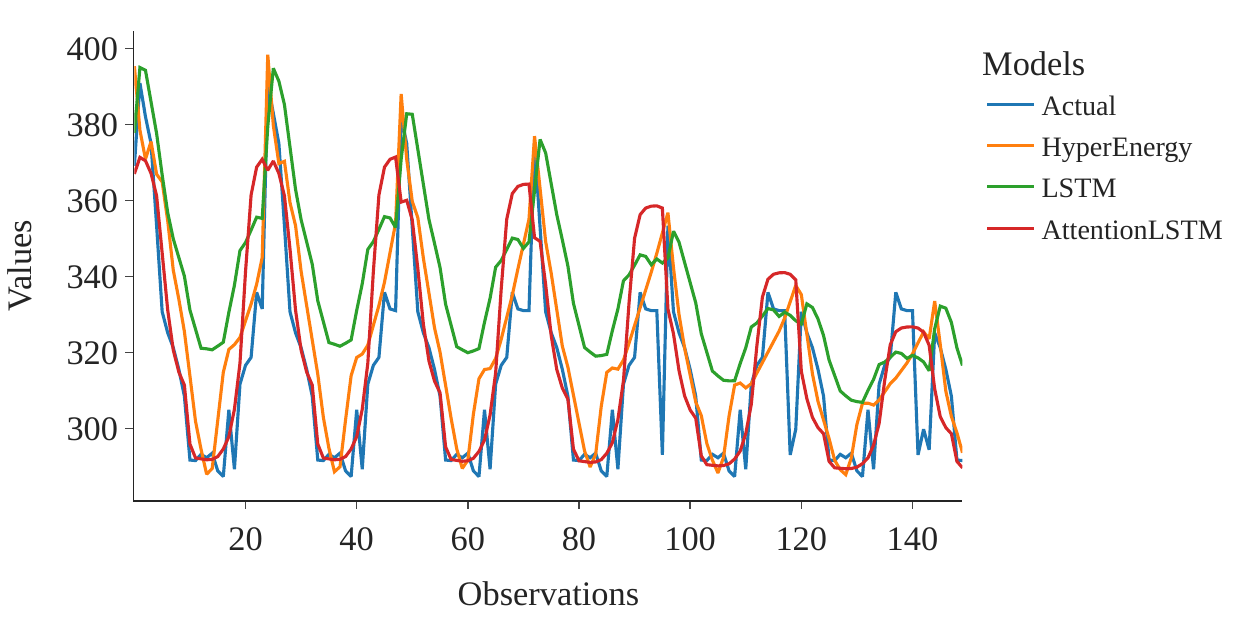}  

    \vspace{-12pt}
    \caption{Student Residence 1: actual versus the predicted value for top four models-- {HyperEnergy}, LSTM, AttentionLSTM, and XGBoost.}
    %\vspace{-10pt}
    \label{fig:line1}

    \centering
    %\vspace{-10pt}
    \includegraphics[trim=0 0 0 0.4cm, clip=true, width=\columnwidth  ]{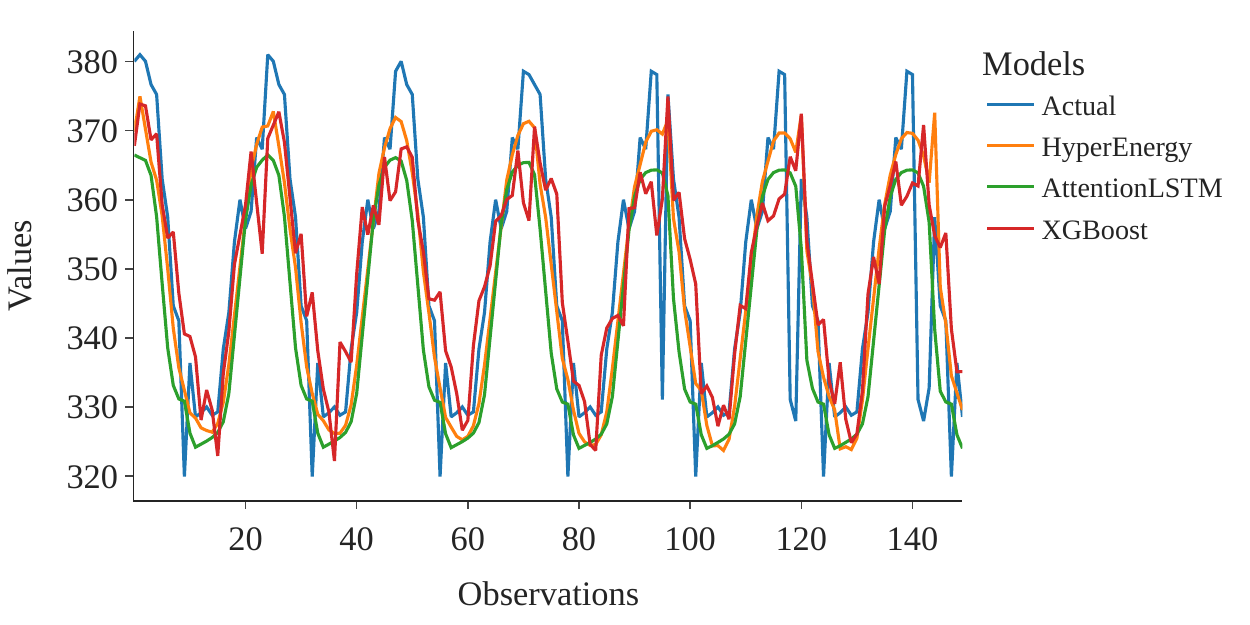}  
     \vspace{-25pt}
     \caption{Student Residence 2: actual versus the predicted value for top four models-- {HyperEnergy}, LSTM, AttentionLSTM, and XGBoost.}
    
    \label{fig:line2}
    \vspace{-10pt}
\end{figure}

Figures \ref{fig:line1} and \ref{fig:line2} show actual versus predicted {energy consumption} for the top four performing models for Residence 1 and Residence 2, respectively. It can be observed that for both residences, {HyperEnergy} captured the patterns better than the remaining models. {In Figure \ref{fig:line2}, although there is seasonality, daily energy peaks are not precisely captured by any of the algorithms. A possible reason for this is that the time period selected for this illustration had slightly different peaks than the training data. Nevertheless, as observed from Table \ref{tab:performance_metrics_residences}, {HyperEnergy} achieves better results than other approaches.}

\subsection{Individual Houses}

\begin{table*}[!h]
\centering
%\caption{Performance Comparison for the Individual households}
\caption{Performance Comparison for the Individual Households {(Testing Days: House 1: 225 days, House 2, 3, and 4: 43 days)}}

\label{tab:performance_metrics_houses}
\begin{tabular}
{@{\hspace{1.2mm}}l@{\hspace{1.2mm}}|@{\hspace{1.2mm}}r@{\hspace{1.2mm}}r@{\hspace{1.2mm}}r@{\hspace{2mm}}|r@{\hspace{1.2mm}}r@{\hspace{1.2mm}}r@{\hspace{2mm}}|r@{\hspace{1.2mm}}r@{\hspace{1.2mm}}r@{\hspace{2mm}}|r@{\hspace{1.2mm}}r@{\hspace{1.2mm}}r@{\hspace{1.2mm}}r@{\hspace{1.2mm}}r@{\hspace{2mm}}r}
\toprule
\multirow{2}{*}{\textbf{Model}} & \multicolumn{3}{c|}{\textbf{House 1}} & \multicolumn{3}{c|}{\textbf{House 2}}  & \multicolumn{3}{c|}{\textbf{House 3 (EV)}} & \multicolumn{3}{c}{\textbf{House 4}} \\
\cmidrule{2-13}
& \textbf{MAE} & \textbf{RMSE} & \textbf{SMAPE}  & \textbf{MAE} & \textbf{RMSE} & \textbf{SMAPE}& \textbf{MAE} & \textbf{RMSE} & \textbf{SMAPE}  & \textbf{MAE} & \textbf{RMSE} & \textbf{SMAPE} \\
\midrule
MLP & 0.47 & 0.52 & 55.40\% & 0.65 & 0.73 & 33.03\% & 0.57 & 0.64 & 45.57\% & 0.32 & 0.45 & 44.05\%  \\ 
NBEATS  & 0.42 & 0.54& 56.32\% & 0.56 & 0.69 & 29.70\% & 0.50 & 0.61 & 43.74\% & 0.38 & 0.46 & 50.86\%\\ 
ARFNN & 0.41 & 0.50 & 50.82\% & 0.53 & 0.63 & 34.60\% & 0.44 & 0.55 & 37.04\% & 0.31 & 0.38 & 44.20\%  \\ 
TempConv & 0.44 & 0.50 & 52.76\% & 0.62 & 0.73 & 33.69\% & 0.49 & 0.56 & 40.80\% & 0.34 & 0.41 & 53.87\% \\ 
XGboost & 0.43 & 0.49 & 52.10\% & 
0.71 & 0.87 & 35.67\% & 0.44 & 0.52 & 37.22\% & 0.31 & 0.37 & 41.37\%  \\ 
RNN & 0.40 & 0.53 & 50.07\% & 0.70 & 0.85 & 35.58\% & 0.45 & 0.66 & 40.39\% & 0.33 & 0.39 & 43.56\% \\ 
GRU & 0.42 & 0.47 & 51.96\% & 0.68 & 0.82 & 34.30\% & 0.38 & 0.52 & 32.03\% & 0.32 & 0.37 & 43.76\% \\ 
LSTM & 0.39 & 0.49 & 48.28\% & 0.63 & 0.82 & 32.53\% &  0.37 & 0.53 & 31.71\% & 0.28 & 0.36 & 39.75\% \\ 
AttentionLSTM & 0.40 & 0.52 & 50.01\% & 0.63 & 0.81 & 32.67\% & 0.38 & 0.51 & 32.59\% & 0.31 & 0.37 & 42.95\% \\ 
Transformer & 0.35 & 0.46 & 45.10\%  & 0.49 & 0.60 & 28.03\% & 0.44 & 0.52 & 37.49\%  & 0.29 & 0.36 & 40.05\% \\ 
\textbf{{HyperEnergy}} & \textbf{0.29} & \textbf{0.38} & \textbf{37.47\%} & \textbf{0.43} & \textbf{0.55} & \textbf{21.03\%} & 
\textbf{0.35} & \textbf{0.53} & \textbf{29.95\%} & 
\textbf{0.27} & \textbf{0.35} & \textbf{37.62\%}
\\ \bottomrule
\end{tabular}
%\vspace{-7pt}
\end{table*}

While the two residencies share some similarities and noticeable patterns due to similarities in students' routines, the four homes exhibit a wide diversity of {energy consumption} patterns as seen from Figures \ref{fig:h1}, \ref{fig:h2}, \ref{fig:h3}, and \ref{fig:h4}. As there is a large randomness component present in these datasets, it is expected the overall accuracy will be lower than that achieved for the residences.

The results for the four houses in terms of the metrics, MAE, RMSE, and SMAPE, are presented in Table \ref{tab:performance_metrics_houses}. For example, {HyperEnergy} achieves SMAPE of 37.47\% which is much higher than SMAPE for student residences. Nevertheless, for House 1, {HyperEnergy} achieves better results than the remaining ten algorithms in therms of all three metrics. Compared to the powerful transformer with SMAPE of 45.10\%, {HyperEnergy} SMAPE of 37.47\% is an improvement of around 8\%. Other approaches achieve weaker performance than {HyperEnergy} and the transformer across most metrics.

\begin{figure}[!t]
    \vspace{-8pt}
    \centering
    \includegraphics[width=0.98\columnwidth]{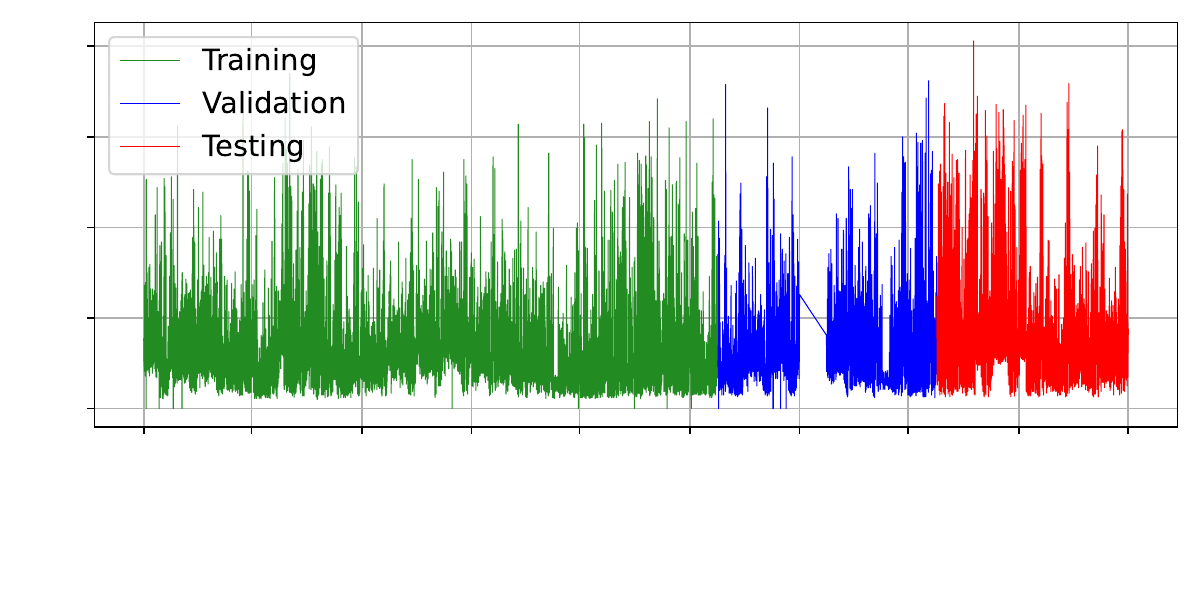}  
    \vspace{-20pt}
    \caption {House 1 {energy consumption}}
    \label{fig:h1}

    \centering
    \vspace{2pt}
    \includegraphics[width=0.98\columnwidth]{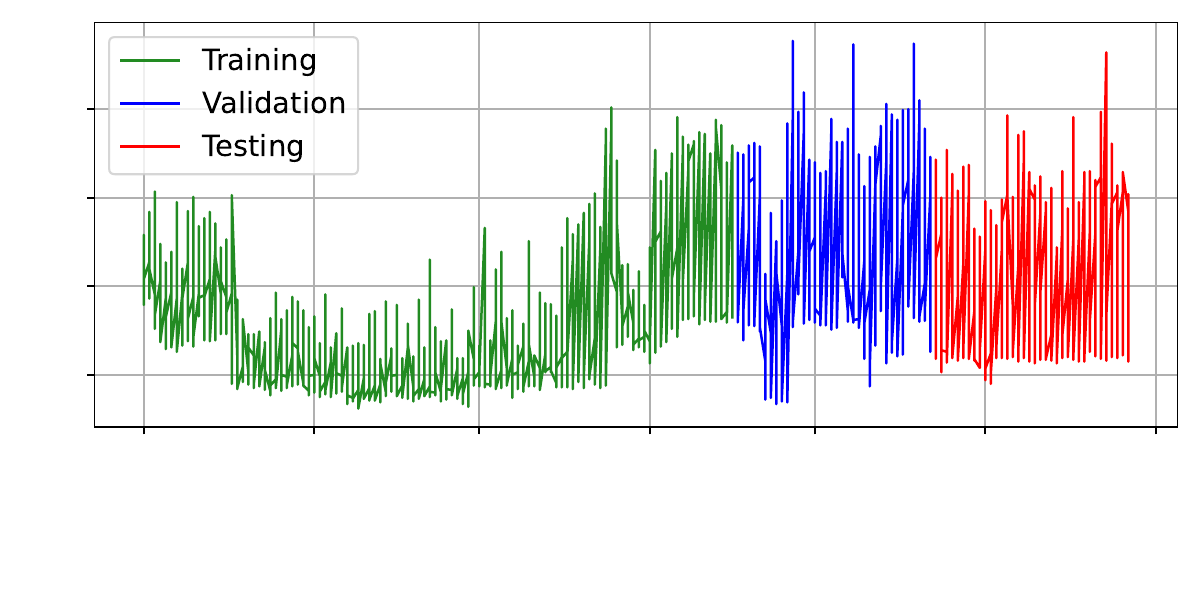}  
    \vspace{-20pt}
    \caption {House 2 {energy consumption}}
    \label{fig:h2}

    \centering
    \vspace{2pt}
    \includegraphics[width=0.98\columnwidth]{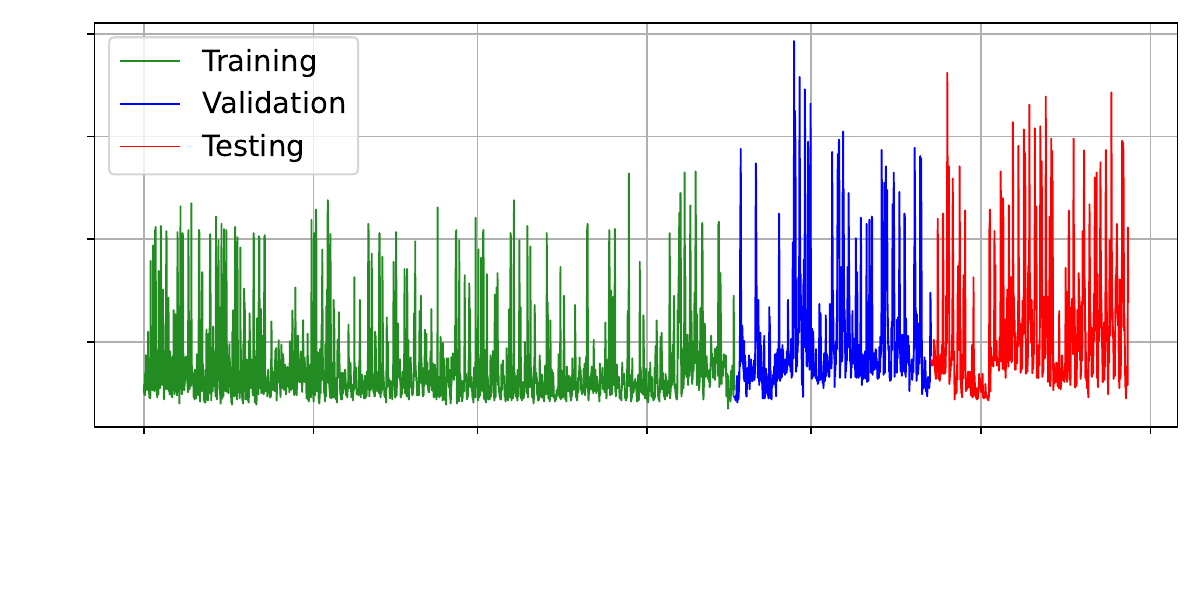}  

    \vspace{-17pt}
    \caption {{House 3 (with electric vehicle) {energy consumption}}}
    \label{fig:h3}

    \centering
    \vspace{2pt}
    \includegraphics[width=0.98\columnwidth]{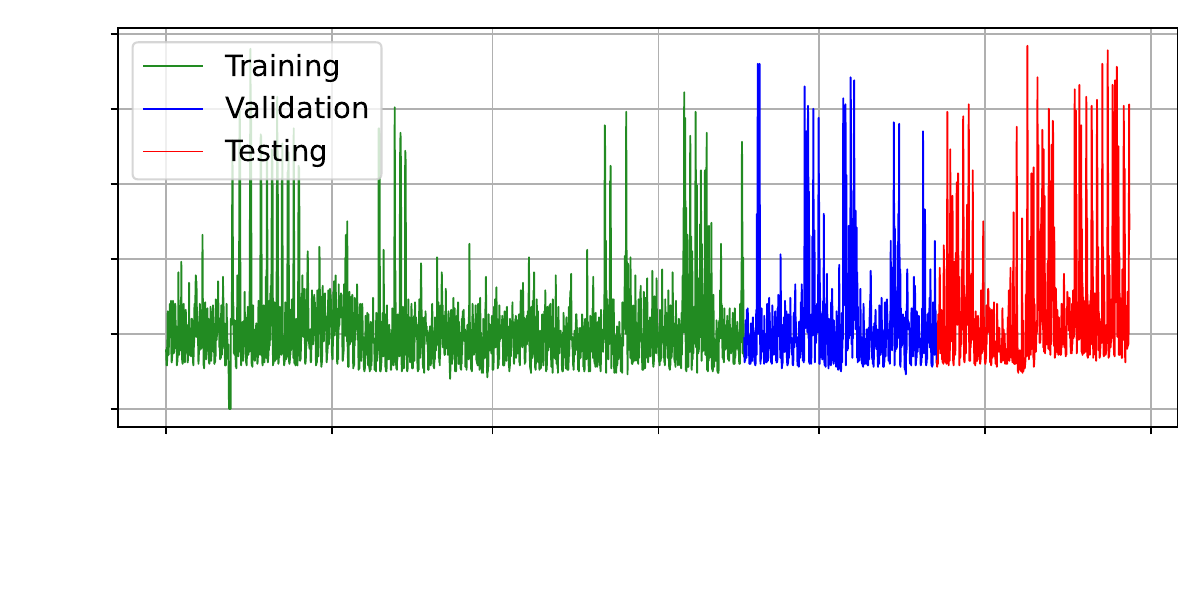}  

    \vspace{-17pt}
    \caption {House 4 (townhouse) {energy consumption}}
    \label{fig:h4}
    \vspace{-8pt}
\end{figure}

For House 2, as seen from Figure \ref{fig:h2}, the {energy consumption} is lower in April and May, while after June there is a level shift and large spike variations, but in test data, there are no trends and seasonality but only sudden spikes and drops. As depicted in Table \ref{tab:performance_metrics_houses}, {HyperEnergy} once again showed superior performance in terms of all three metrics, by recording a MAE of 0.43, RMSE of 0.55, and SMAPE of 21.03\%. The transformer lagged behind with an MAE of 0.49, RMSE of 0.60, and SMAPE of 28.03\%. In contrast, all other models recorded SMAPE values exceeding 30\%. These numbers demonstrate the ability of the {HyperEnergy} to capture complex data patterns. Note that SMAPE for House 2 is lower than for House 1 for all models showing increased predictability for House 2. On the other hand, MAE and RMSE cannot be compared across houses as they are scale-dependent metrics. 

With House 3, we extend the evaluation to a house with an Electric Vehicle (EV) which is expected to cause charging spikes. These sudden spikes, together with level shifts can be observed in Figure \ref{fig:h3}. {HyperEnergy} achieves the lowest errors across all metrics, with MAE of 0.35, RMSE of 0.53, and SMAPE of 29.95\%. Several other models such as GRU, LSTM, and AttentionLSTM also show commendable performance, with SMAPE values hovering around  31\%. 

House 4 again increases the diversity of consumers by considering a townhouse instead of a standalone house. Again, sudden spikes are highly present in data as seen from Figure 9. Yet again, the HyperForecasting showed superior performance in terms of all three metrics. LSTM, AttentionLSTM, Transformer, and XGboost follow closely with SMAPE values close to 40\%. While 37.62\% achieved by HyperForecasting for House 4 is much higher than that achieved for residences, House 2, and House 3, it is still better than the other considered models.

\subsection{{Industrial and Commercial Buildings}}

\begin{table*}[!t]
\centering
\caption{{Performance Comparison for Industrial and Commercial Buildings} {(Testing Days: 146 days)}}
\label{tab:performance_metrics_cbuildings}
%\color{blue}
\begin{tabular} {@{\hspace{1.2mm}}l@{\hspace{1.2mm}}|@{\hspace{1.2mm}}r@{\hspace{1.2mm}}r@{\hspace{1.2mm}}r@{\hspace{2mm}}|r@{\hspace{1.2mm}}r@{\hspace{1.2mm}}r@{\hspace{2mm}}|r@{\hspace{1.2mm}}r@{\hspace{1.2mm}}r@{\hspace{2mm}}|r@{\hspace{1.2mm}}r@{\hspace{1.2mm}}r@{\hspace{1.2mm}}r@{\hspace{1.2mm}}r@{\hspace{2mm}}r}
\toprule
\multirow{2}{*}{\textbf{Model}} & \multicolumn{3}{c|}{\textbf{Manufacturing}} & \multicolumn{3}{c|}{\textbf{Medical Clinic}}  & \multicolumn{3}{c|}{\textbf{Retail Store}} & \multicolumn{3}{c}{\textbf{Office}} \\
\cmidrule{2-13}
& \textbf{MAE} & \textbf{RMSE} & \textbf{SMAPE}  & \textbf{MAE} & \textbf{RMSE} & \textbf{SMAPE}& \textbf{MAE} & \textbf{RMSE} & \textbf{SMAPE}  & \textbf{MAE} & \textbf{RMSE} & \textbf{SMAPE} \\
\midrule
MLP & 54.12 & 63.00 & 8.53\% & 11.62 & 14.05 & 4.62\% & 44.12 & 49.36 & 13.78\% & 27.95 & 30.55 & 5.43\% \\ 
NBEATS & 52.14 & 61.99 & 8.16\% & 8.60 & 10.48 & 3.51\% & 50.79 & 64.33 & 16.82\% & 34.12 & 42.20 & 7.06\% \\ 
ARFNN & 55.59 & 65.22 & 8.70\% & 11.85 & 14.59 & 4.75\% & 23.96 & 28.67 & 7.57\% & 20.07 & 23.11 & 3.98\% \\ 
Conv1*1 & 57.36 & 66.55 & 8.76\% & 9.12 & 10.92 & 3.57\% & 34.35 & 38.75 & 10.23\% & 15.08 & 18.20 & 3.07\% \\ 
XGBoost & 73.26 & 87.07 & 11.36\% & 16.62 & 19.22 & 6.61\% & 34.51 & 39.87 & 10.61\% & 23.71 & 27.54 & 4.55\% \\ 
RNN & 74.15 & 88.71 & 11.66\% & 19.44 & 23.65 & 7.88\% & 33.36 & 38.57 & 10.10\% & 20.90 & 24.64 & 4.18\% \\ 

GRU & 55.59 & 65.22 & 8.70\% & 11.85 & 14.59 & 4.75\% & 23.96 & 28.67 & 7.57\% & 20.07 & 23.11 & 3.98\% \\ 

LSTM  & 48.45 & 56.87 & 7.67\%& 10.28 & 12.91 & 4.05\% & 22.25 & 26.67 & 7.56\% & 19.13 & 22.54 & 3.77\% \\ 
AttentionLSTM & 50.40 & 59.65 & 7.89\% & 9.07 & 10.54 & 3.60\% & 25.36 & 31.52 & 8.30\% & \textbf{11.70} & \textbf{14.50} & \textbf{2.50}\% \\ 
Transformer & 43.50 & \textbf{51.87} & 7.56\% & 9.76 & 10.70 & 3.65\% & \textbf{20.48} & \textbf{24.48} & 7.67\% & 15.34 & 18.23 & 3.14\% \\ 
\textbf{{HyperEnergy}} & \textbf{40.34} & {52.16} & \textbf{6.29\%} & \textbf{5.98} & \textbf{9.36} & \textbf{2.42\%} & {22.22} & {28.13} & \textbf{7.38\%} & 13.33 & 17.16 & 2.82\% \\ 
\bottomrule
\end{tabular}
\vspace{-7pt}
\end{table*}
\color{black}

Table \ref{tab:performance_metrics_cbuildings} presents the results for four types of industrial and commercial buildings: a manufacturing building, a medical clinic, a retail store, and an office building. For these buildings, SMAPE values are much lower than those observed for student residences (Table \ref{tab:performance_metrics_residences}) and individual homes (Table \ref{tab:performance_metrics_houses}): most algorithms archived SMAPE values under 12\%. This can be explained by these buildings having more consistent energy use patterns due to regular working hours and activities.

 For the manufacturing building, {HyperEnergy} achieved the best MAE and SMAPE values, while the transformer achieved slightly better RMSE. For the medical clinic, the transformer achieved the best results in terms of all three metrics. For the retail store, MAE and RMSE were better for the transformer than for {HyperEnergy}, and finally, for the office building, the Attention LSTM achieved the best results in terms of all three metrics.

 Looking across all four buildings, the three top models -- Attention LSTM, Transformer, and our HyperEnergy --achieved similarly high levels of accuracy in their predictions, with {HyperEnergy} demonstrating a slight advantage in most metrics. Combined with findings for student residences and homes, this demonstrates that our HyperEnergy is a suitable consumer energy forecasting technique for a diverse range of consumers.

\subsection{Ablation Studies} 
The learnable adaptive kernel in {HyperEnergy} plays a crucial role in modeling complex energy patterns and enhancing the prediction. This section examines the significance of this component as well as the impact of kernel type. The results are presented for Residences and House 2, while the remaining datasets are omitted for brevity.
\subsubsection{Study 1: With and Without Learnable Adaptive Kernel}
As seen from Table \ref{tab:ablation1}, for Residence 2, {HyperEnergy} with the learnable adaptive kernel achieved SMAPE of 6.70\% which is a noticeable improvement compared to the version without the kernel which obtains SMAPE of 7.76\%. It is also important to note that the LSTM alone achieved SMAPE of 9.29\% which indicates that the inclusion of a hypernetwork in our approach contributes to error reduction while the addition of kernels further reduces the error. Figure \ref{fig:ablationstudy1} compares the predictions obtained for Residence 2 with and without kernel to the actual values: it is noticeable that with the kernel, predictions better follow the actual values.

To extend the examination, the same analysis but for House 2 is depicted in Table \ref{tab:ablation2}. Again, the learnable adaptive kernel improves predictions by reducing SMAPE from 24.47\% to 21.03\%. Also {HyperEnergy} achieves much better accuracy than the LSTM model. From Figure \ref{fig:ablationstudy2} it can be observed that predictions with the kernel better follow the actual values than those without the kernel. 

Overall, {HyperEnergy} without the kernel archives better performance than the standalone LSTM justifying the adaptation of hypernetworks. Moreover, the learnable adaptive kernel further enhances predictive performance.

\begin{table}[t!]
\centering
\caption{Ablation Study 1, Residence 2: Metrics for LSTM and {HyperEnergy} with and without the Adaptive Kernel.}
\label{tab:ablation1}
\begin{tabular}{ p{3.5cm}p{1cm}p{1cm}p{1cm}  }
\toprule
\textbf{Model} & \textbf{MAE} & \textbf{RMSE} & \textbf{SMAPE} \\
\midrule
LSTM & 21.76 & 30.33 & 9.29\% \\ 
{HyperEnergy} without & & &\\
Learnable Adaptive Kernel & 19.04 & 26.65 & 7.76\% \\ 
\textbf{{HyperEnergy} with} & & &\\ \textbf{Learnable Adaptive Kernel} & \textbf{16.28} & \textbf{24.59} & \textbf{6.70\%} \\ 
\bottomrule
\end{tabular}
%\vspace{10pt}
\end{table}

\begin{figure}[t]
    \centering
    \vspace{-5pt}
    \includegraphics[trim=0 0 0 2.5cm, clip=true, width=\columnwidth]{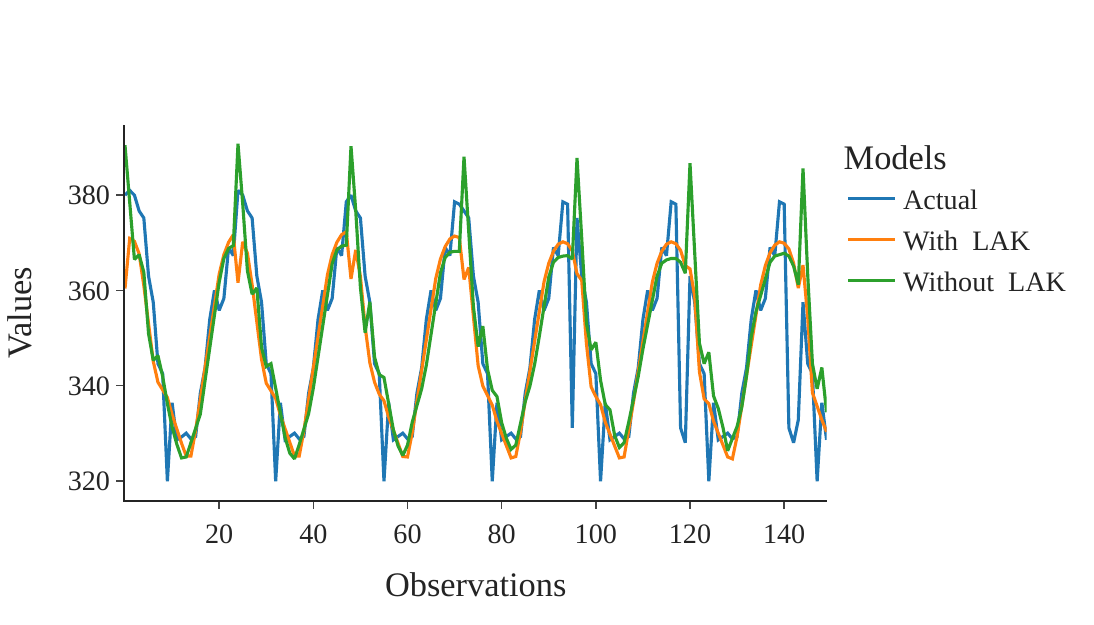} 
    \vspace{-25pt}
    \caption{Ablation Study 1, Residence 2: {HyperEnergy} prediction with and without the learnable adaptive kernel compared to actual values.}
    \vspace{-5pt}
    \label{fig:ablationstudy1}
\end{figure}

\begin{table}[t]
\centering
\caption{Ablation Study 1, House 2: Metrics for LSTM and {HyperEnergy} with and without the Adaptive Kernel.}
\label{tab:ablation2}
\begin{tabular}{ p{3.5cm}p{1cm}p{1cm}p{1cm}  }
\toprule
\textbf{Model} & \textbf{MAE} & \textbf{RMSE} & \textbf{SMAPE} \\
\midrule
LSTM & 0.63 & 0.82 & 32.53\% \\ 
{HyperEnergy} without &&&\\
Learnable Adaptive Kernel & 0.49 & 0.62 & 24.47\% \\ 
\textbf{{HyperEnergy} with} &&&\\
\textbf{Learnable Adaptive Kernel} & \textbf{0.43} & \textbf{0.55} & \textbf{21.03\%} \\ \bottomrule
\end{tabular}
\end{table}

\begin{figure}[t]
    \centering
    
    \includegraphics[trim=0 0 0 2.5cm, clip=true, width=\columnwidth]{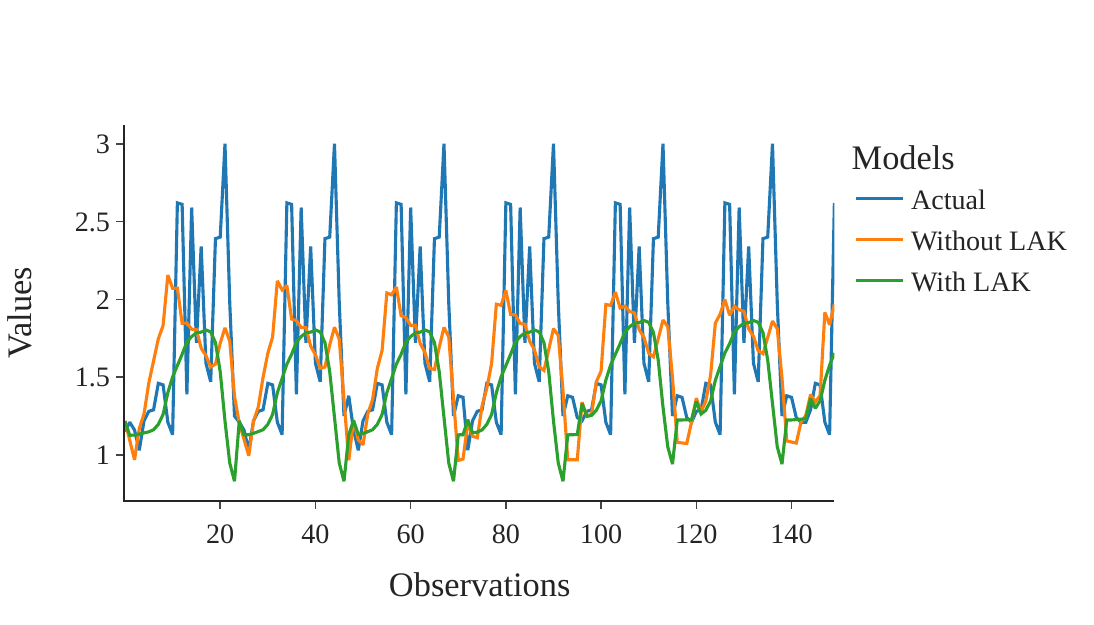}  
    \vspace{-25pt}
    \caption{Ablation Study 1, House 2: {HyperEnergy} predictions with and without the learnable adaptive kernel compared to actual values.}
    \vspace{-10pt}
    \label{fig:ablationstudy2}
\end{figure}

\subsubsection{Study 2: With Learnable and Traditional Kernels}

{We extended the evaluation to demonstrate the need for learnable kernels and the necessity to merge polynomial and RBF kernels. Table \ref{tab:ablation3} first compares learnable and traditional kernels for RBF and the polynomial kernel, respectively, and then examines the combination of traditional and learnable kernels. {HyperEnergy} with a learnable RBF outperformed the traditional RBF, achieving a SMAPE of 8.76\% compared to the traditional RBF's 9.04\%. Similarly, the learnable polynomial kernel achieved a 9.46\% SMAPE compared to the traditional polynomial kernel's 10.58\%. In both cases, the RBF and the polynomial kernel, the learnable version performed better than the traditional one.}

{Comparing the combined traditional kernels (the second-to-last row in Table \ref{tab:ablation3}) with the combined learnable kernels (the last row in the table), replacing traditional kernels with learnable ones improves performance, reducing SMAPE from 9.31\% to 8.27\%. This demonstrates that learnable kernels improve performance over traditional kernels and shows that combining kernels improves accuracy.}

\begin{table}[t]
\centering
\caption{Ablation Study 2, Residence 1: {HyperEnergy} with the Learnable and Traditional Kernels.}
\label{tab:ablation3}
\begin{tabular}{ p{4.5cm} p{0.8cm}p{0.8cm}p{0.8cm}  }
\toprule
\textbf{Model} & \textbf{MAE} & \textbf{RMSE} & \textbf{SMAPE} \\ \midrule
{{HyperEnergy} (Traditional RBF} & {22.10} & {30.50} & {9.04\%} \\
{{HyperEnergy} (Learnable RBF)} & 21.21 & 29.30 & 8.76\% \\
{{HyperEnergy} (Traditional Polynomial)} & {25.80} & {34.50} & {10.58\%} \\
{{HyperEnergy} (Learnable Polynomial)} & 23.17 & 32.23 & 9.46\% \\
{{HyperEnergy} (Traditional Combined)} & {23.10} & {31.40} & {9.31\%} \\
\textbf{{HyperEnergy} with Learnable} &&& \\
\textbf{Adaptive Kernel} & \textbf{20.02} & \textbf{27.58} & \textbf{8.27\%} \\ \bottomrule
\end{tabular}
\end{table}

\begin{table*}[!h]
\centering
\caption{{Comparison of Computation (in Minutes) of Selected Models}}
\label{tab:computation}
%\color{blue}
\begin{tabular} {@{\hspace{1.2mm}}l@{\hspace{1.2mm}}|@{\hspace{1.2mm}}r@{\hspace{1.2mm}}r@{\hspace{1.2mm}}r@{\hspace{2mm}}|r@{\hspace{1.2mm}}r@{\hspace{1.2mm}}r@{\hspace{2mm}}|r@{\hspace{1.2mm}}r@{\hspace{1.2mm}}r@{\hspace{2mm}}|r@{\hspace{1.2mm}}r@{\hspace{1.2mm}}r@{\hspace{1.2mm}}r@{\hspace{1.2mm}}r@{\hspace{2mm}}r}
\toprule
\multirow{2}{*}{\textbf{Model}} & \multicolumn{3}{c|}{\textbf{Residence 1}} & \multicolumn{3}{c|}{\textbf{House 1}}  & \multicolumn{3}{c|}{\textbf{House 2}} & \multicolumn{3}{c}{\textbf{Industry}} \\
\cmidrule{2-13}
& \textbf{Training} & \textbf{Testing} & \textbf{}  & \textbf{Training} & \textbf{Testing} & \textbf{} & \textbf{Training} & \textbf{Testing} & \textbf{}  & \textbf{Training} & \textbf{Testing} & \textbf{} \\
\midrule
LSTM & 39.9 & 0.13 &  & 26.2 & 0.08 &  & 8.6 & 0.02 &  & 15.2 & 0.04 &  \\ 
GRU & 30.0 & 0.12 &  & 20.0 & 0.06 &  & 7.1 & 0.02 &  & 12.0 & 0.03 &  \\ 
AttentionLSTM & 45.5 & 0.15 &  & 30.8 & 0.10 &  & 10.2 & 0.03 &  & 21.8 & 0.06 &  \\ 
Transformer & 118.3 & 0.28 &  & 98.0 & 0.20 &  & 14.2 & 0.16 &  & 45.2 & 0.08 &  \\ 
NBEATS & 120.4 & 0.29 &  & 98.3 & 0.22 &  & 14.3 & 0.15 &  & 46.8 & 0.08 &  \\ 
{HyperEnergy} & 62.9 & 0.18 &  & 35.9 & 0.12 &  & 9.8 & 0.03 &  & 27.1 & 0.07 &  \\ 
\bottomrule
\end{tabular}
\vspace{-7pt}
\end{table*}
\color{black}

\begin{figure*}[!h]
    \centering
    \includegraphics[width=0.9\textwidth]{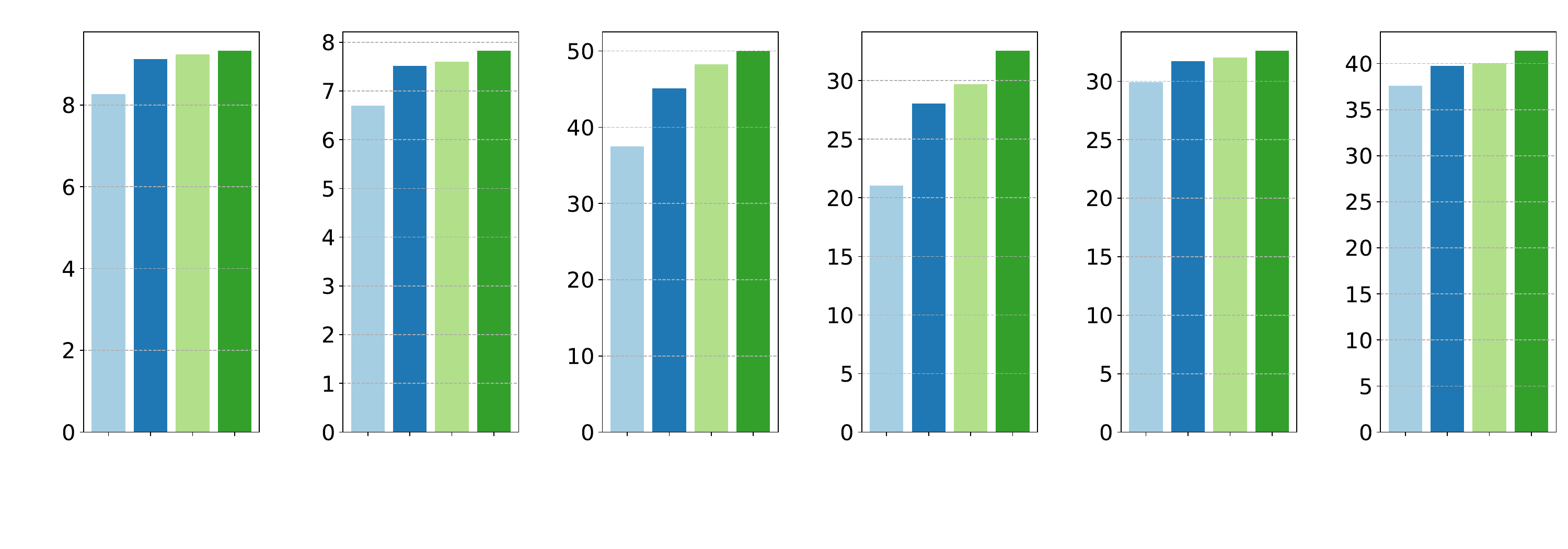} 
    \vspace{-13pt}
    \caption{Comparison of {HyperEnergy} with best three approaches for each consumer. Note that the best approaches are not the same across the consumers, but for all consumers, {HyperEnergy} achieves the lowest SMAPE }
    
    \label{fig:lineoverall}
    \vspace{-10pt}
\end{figure*}

\subsection{Discussion}
In many energy forecasting studies, the LSTM memory mechanism is key to capturing and retaining information from historical usage, which is then used to predict future energy consumption patterns \cite{shaikh2023new}, \cite{bashir2022short}, \cite{kong2017short}. The proposed HyperEnergy transforms and stabilizes this process by predicting optimized weights and biases for the LSTM, helping it not only enhance performance but also maintain consistency across diverse datasets. The inclusion of learnable adaptive kernels, which combine polynomial and RBF kernels, enables the model to capture both gradual and sudden shifts in consumption behavior. The Parameter Integration Module seamlessly assigns weights and biases to the LSTM's internal gates during training, helping LSTM maintain consistency across datasets with varied characteristics and usage behavior. As seen in Figure \ref{fig:lineoverall}, while LSTM alone performs well in cases such as Residence 1, House 3, and House 4, it ranks lower for Residence 2. However, our proposed learning method ensures that HyperEnergy still outperforms all other techniques for all houses and residences, including Residence 2. As demonstrated in our ablation studies (Section V-D), both the hypernetwork and learnable kernels contribute to notable error reduction. These findings demonstrate that HyperEnergy can be effectively applied in scenarios requiring accurate predictions across diverse consumer types.

For the practical applicability of our method, it is crucial to examine both training and inference (prediction) times. Table \ref{tab:computation} compares our approach to top five approaches in terms of training and testing (inference) time. Four buildings were selected, as the remaining buildings have similar numbers of samples and computation times: for example, the two residences have the same number of samples, while Houses 3 and 4 have the same number of samples as House 2. All models were trained and tested on a computer with an AMD Ryzen Threadripper PRO 5955WX processor and an NVIDIA GA102GL RTX A6000 GPU. Note that testing time in Table \ref{tab:computation} is shown for the complete test set which ranges from 43 to 335 days depending on the dataset (20\% of available data). Even though LSTM, GRU, and AttentionLSTM exhibit shorter inference times, our approach achieves superior prediction accuracy, as evidenced by the results presented in Tables \ref{tab:performance_metrics_residences}, \ref{tab:performance_metrics_houses}, and \ref{tab:performance_metrics_cbuildings}, with a maximum inference time of only 0.18 minutes for the complete test set.

The practicality of the proposed approach is further demonstrated through its ability to handle diverse building types without the need for algorithm modification. For new buildings, the algorithm is trained with that building's data: in our experiments, the longest training time was around 120 minutes for Residence 1 (Table \ref{tab:computation}). In contrast, existing studies fail to demonstrate the broad applicability of their solutions across diverse consumer profiles, thereby limiting their practical applications and real-world utility. Overall, {HyperEnergy} achieved better prediction accuracy than other techniques across diverse datasets, demonstrating its ability to capture complex dependencies and its applicability across different consumer types.

Once the model is trained, it is used for inference. In our experiments, the testing period was 20\% of the training data, resulting in 43 to 335 days depending on the dataset; during this period, the model remained unchanged. However, changes in energy consumption patterns may lead to model degradation, and at that time, the model could benefit from retraining.

As demonstrated through extensive experiments, our approach generally outperforms other techniques. However, there are occasional short periods where other techniques achieve slightly better predictions. For instance, during an early winter weekend in Residence 1, the transformer maintained an SMAPE below 5\%, while our method exceeded 7\% between 21:00 and 07:00. Nevertheless, our approach was overall better for that residence.

While our approach overall performed the best, the achieved prediction accuracy varied greatly across the datasets. Student residences, as observed from Figures \ref{fig:essex_energy} to \ref{fig:line2} show clear seasonal patterns due to uniform routines, which our method captured well, achieving an SMAPE of under 9\%. In contrast, individual houses exhibited high energy variance without strong seasonal patterns, Figures \ref{fig:h1} to \ref{fig:h4}, resulting in lower accuracy for all models, with our model achieving an SMAPE between 21\% and 38\%, Table \ref{tab:performance_metrics_houses}. Finally, for industrial and commercial buildings, as shown in Table \ref{tab:performance_metrics_cbuildings}, SMAPE values were lower: manufacturing 6.29\%, medical clinic 2.42\%, retail store 7.38\%, and office building 2.82\%. Such low errors can be explained by stable patterns caused by their operation schedules.

ML models are highly suitable for consumer-level energy consumption forecasting, resulting in their successful deployment in numerous consumer-level energy forecasting applications \cite{ gong2021peak, kong2017short, zhang2020deep, lin2022hybrid}. The ML models efficiently learn patterns from historical energy usage data of each building, enabling them to deliver accurate predictions that enhance their applicability. Moreover, ML models generally outperform traditional statistical methods such as autoregressive and moving average models in handling non-linear and fluctuating consumption behaviors \cite{klyuev2022methods}. Our proposed method HyperEnergy is also an ML-based deep learning network that specializes in keeping consistent performance across diverse consumer types. This novel approach addresses the limitations of conventional machine learning models by continuously optimizing LSTM parameters through the hypernetwork and learnable kernels, allowing for error reduction and consistency, even across diverse consumer energy profiles. As shown in our extensive experiments, HyperEnergy not only outperforms state-of-the-art models but also adapts effectively to both gradual and sudden shifts in energy consumption, ensuring robust and reliable forecasting for a wide range of consumer behaviors.

Overall trends can be observed throughout case studies. The performance of ML models tends to vary greatly across buildings due to differences in the variability of energy consumption patterns among buildings. While SMAPE is under 10\% for residences and under 8\% for industrial and commercial buildings, it increases greatly for individual houses, often exceeding 30\%. Although the error of our approach varies, increasing for homes in comparison to residences, our approach outperforms other models in most cases.

\section{Conclusion} 
\label{sec:conclusion}
This paper proposed {HyperEnergy}, a {consumer energy forecasting} approach founded on the hypernetworks suitable for forecasting energy across diverse categories of consumers. Kernelized hypernetwork is integrated with the LSTM model to improve captioning complex consumption patterns by updating LSTM parameters through a meta-network. The forecasting is further improved through the learnable kernel which integrates adapted polynomial and RBF kernels while changing the impact of kernels though the learning process.

The evaluation was conducted on diverse consumers: two student residencies and four individual households including two detached homes, one home with EV charging, and one townhouse. Across nine datasets, the proposed {HyperEnergy} outperformed 10 other forecasting approaches including state-of-the-art deep learning models such as LSTM, AttentionLSTM, and transformer. The ablation studies demonstrated that including the hypernetwork for setting parameters of the primary network improved the forecasting accuracy. The adaptable kernel improved accuracy although to a lesser extent than the hypernetwork itself.
Figure \ref{fig:lineoverall} compares {HyperEnergy} with the top three other approaches for each of the ten datasets. Note that to top three performers besides {HyperEnergy} differ among the consumers.  
While the accuracy varies across consumers, {HyperEnergy} achieved the best results for each consumer.

Future work will investigate reusing already trained models between similar consumers through transfer learning. Moreover, the approach will be evaluated on additional datasets.

\bibliographystyle{IEEEtran}
\bibliography{ref.bib}

\begin{thebibliography}{10}
\providecommand{\url}[1]{#1}
\csname url@samestyle\endcsname
\providecommand{\newblock}{\relax}
\providecommand{\bibinfo}[2]{#2}
\providecommand{\BIBentrySTDinterwordspacing}{\spaceskip=0pt\relax}
\providecommand{\BIBentryALTinterwordstretchfactor}{4}
\providecommand{\BIBentryALTinterwordspacing}{\spaceskip=\fontdimen2\font plus
\BIBentryALTinterwordstretchfactor\fontdimen3\font minus \fontdimen4\font\relax}
\providecommand{\BIBforeignlanguage}[2]{{%
\expandafter\ifx\csname l@#1\endcsname\relax
\typeout{** WARNING: IEEEtran.bst: No hyphenation pattern has been}%
\typeout{** loaded for the language `#1'. Using the pattern for}%
\typeout{** the default language instead.}%
\else
\language=\csname l@#1\endcsname
\fi
#2}}
\providecommand{\BIBdecl}{\relax}
\BIBdecl

\bibitem{eia2023}
\BIBentryALTinterwordspacing
{U.S. Energy Information Administration}, ``Eia projects nearly 50\% increase in world energy usage by 2050,'' 2023. [Online]. Available: \url{https://www.eia.gov/todayinenergy/detail.php?id=42342}
\BIBentrySTDinterwordspacing

\bibitem{eu2022}
\BIBentryALTinterwordspacing
{European Climate, Energy and Environment}. (2022) 2030 climate \& energy framework. [Online]. Available: \url{https://climate.ec.europa.eu/eu-action/climate-strategies-targets/2030-climate-energy-framework_en}
\BIBentrySTDinterwordspacing

\bibitem{doe2012}
\BIBentryALTinterwordspacing
{U.S. Department of Energy}, ``Reliability improvements from the application of distribution automation technologies,'' Tech. Rep.~3, 2012. [Online]. Available: \url{https://www.energy.gov/sites/prod/files/2016/10/f33/Distribution_Reliability_Report_-_Final_Dec_2012.pdf}
\BIBentrySTDinterwordspacing

\bibitem{chen2021application}
T.~Chen, G.~Chen, W.~Chen, S.~Hou, Y.~Zheng, and H.~He, ``Application of decoupled {ARMA} model to modal identification of linear time-varying system based on the {ICA} and assumption of “short-time linearly varying”,'' \emph{Journal of Sound and Vibration}, 2021.

\bibitem{aslam2021survey}
S.~Aslam, H.~Herodotou, S.~M. Mohsin, N.~Javaid, N.~Ashraf, and S.~Aslam, ``A survey on deep learning methods for power load and renewable energy forecasting in smart microgrids,'' \emph{Renewable and Sustainable Energy Reviews}, 2021.

\bibitem{lheureux2022transformer}
A.~L’Heureux, K.~Grolinger, and M.~Capretz, ``Transformer-based model for electrical load forecasting,'' \emph{Energies}, 2022.

\bibitem{cai2019day}
M.~Cai, M.~Pipattanasomporn, and S.~Rahman, ``Day-ahead building-level load forecasts using deep learning vs. traditional time-series techniques,'' \emph{Applied Energy}, 2019.

\bibitem{das2023long}
A.~Das, W.~Kong, A.~Leach, R.~Sen, and R.~Yu, ``Long-term forecasting with {TiDE}: Time-series dense encoder,'' \emph{arXiv preprint arXiv:2304.08424}, 2023.

\bibitem{alqushaibi2020review}
A.~Alqushaibi, S.~J. Abdulkadir, H.~M. Rais, and Q.~Al-Tashi, ``A review of weight optimization techniques in recurrent neural networks,'' in \emph{IEEE International Conf. on Computational Intelligence}, 2020.

\bibitem{gong2021peak}
H.~Gong, V.~Rallabandi, M.~L. McIntyre, E.~Hossain, and D.~M. Ionel, ``Peak reduction and long term load forecasting for large residential communities including smart homes with energy storage,'' \emph{IEEE Access}, 2021.

\bibitem{rezaei2020optimal}
E.~Rezaei and H.~Dagdougui, ``Optimal real-time energy management in apartment building integrating microgrid with multizone hvac control,'' \emph{IEEE Transactions on Industrial Informatics}, 2020.

\bibitem{kong2017short}
W.~Kong, Z.~Y. Dong, Y.~Jia, D.~J. Hill, Y.~Xu, and Y.~Zhang, ``Short-term residential load forecasting based on {LSTM} recurrent neural network,'' \emph{IEEE Transactions on Smart Grid}, 2017.

\bibitem{zhang2020deep}
X.~Zhang, K.~W. Chan, H.~Li, H.~Wang, J.~Qiu, and G.~Wang, ``Deep-learning-based probabilistic forecasting of electric vehicle charging load with a novel queuing model,'' \emph{IEEE Transactions on Cybernetics}, 2020.

\bibitem{zhang2018forecasting}
X.~Zhang, K.~Grolinger, and M.~A. Capretz, ``Forecasting residential energy consumption using support vector regressions,'' in \emph{Proceedings of the IEEE International Conference on Machine Learning and Applications, Orlando, FL, USA}, 2018.

\bibitem{madhukumar2022regression}
M.~Madhukumar, A.~Sebastian, X.~Liang, M.~Jamil, and M.~N. S.~K. Shabbir, ``Regression model-based short-term load forecasting for university campus load,'' \emph{IEEE Access}, 2022.

\bibitem{ni2024light}
C.~Ni, H.~Huang, and P.~e.~a. Cui, ``Light gradient boosting machine (lightgbm) to forecasting data and assisting the defrosting strategy design of refrigerators,'' \emph{International Journal of Refrigeration}, 2024.

\bibitem{oreshkin2021n}
B.~N. Oreshkin, G.~Dudek, , and E.~Turkina, ``N-beats neural network for mid-term electricity load forecasting,'' \emph{Applied Energy}, 2021.

\bibitem{shaikh2023new}
A.~K. Shaikh, A.~Nazir, and K.~et~al., ``A new approach to seasonal energy consumption forecasting using temporal convolutional networks,'' \emph{Results in Engineering}, 2023.

\bibitem{yamak2019comparison}
P.~T. Yamak, L.~Yujian, and P.~K. Gadosey, ``A comparison between {ARIMA}, {LSTM}, and {GRU} for time series forecasting,'' in \emph{2nd international conference on algorithms, computing and artificial intelligence}, 2019.

\bibitem{skala2023interval}
R.~Skala, M.~A.~T. Elgalhud, K.~Grolinger, and S.~Mir, ``Interval load forecasting for individual households in the presence of electric vehicle charging,'' \emph{Energies}, 2023.

\bibitem{vaswani2017attention}
A.~Vaswani, N.~Shazeer, N.~Parmar, J.~Uszkoreit, L.~Jones, A.~N. Gomez, {\L}.~Kaiser, and I.~Polosukhin, ``Attention is all you need,'' \emph{Advances in neural information processing systems}, vol.~30, 2017.

\bibitem{li2023short}
S.~Li, X.~Kong, L.~Yue, C.~Liu, M.~A. Khan, Z.~Yang, and H.~Zhang, ``Short-term electrical load forecasting using hybrid model of manta ray foraging optimization and support vector regression,'' \emph{Journal of Cleaner Production}, 2023.

\bibitem{li2023novel}
Y.~Li, N.~Zhu, and Y.~Hou, ``A novel hybrid model for building heat load forecasting based on multivariate empirical modal decomposition,'' \emph{Building and Environment}, 2023.

\bibitem{triebe2019ar}
O.~Triebe, N.~Laptev, and R.~Rajagopal, ``{AR-Net}: A simple auto-regressive neural network for time-series,'' \emph{arXiv preprint arXiv:1911.12436}, 2019.

\bibitem{lin2022hybrid}
X.~Lin, R.~Zamora, C.~A. Baguley, and A.~K. Srivastava, ``A hybrid short-term load forecasting approach for individual residential customer,'' \emph{IEEE Transactions on Power Delivery}, 2022.

\bibitem{hoopes2021hypermorph}
A.~Hoopes, M.~Hoffmann, B.~Fischl, J.~Guttag, and A.~V. Dalca, ``Hypermorph: Amortized hyperparameter learning for image registration,'' in \emph{International Conference on Information Processing in Medical Imaging}, 2021.

\bibitem{schmidhuber1993self}
J.~Schmidhuber, ``A ‘self-referential’weight matrix,'' in \emph{International Conference on Artificial Neural Networks}, 1993.

\bibitem{li2020dhp}
Y.~Li, S.~Gu, K.~Zhang, L.~Van~Gool, and R.~Timofte, ``Dhp: Differentiable meta pruning via hypernetworks,'' in \emph{16th European Conference on Computer Vision}, 2020.

\bibitem{meyerson2019modular}
E.~Meyerson and R.~Miikkulainen, ``Modular universal reparameterization: Deep multi-task learning across diverse domains,'' \emph{Advances in Neural Information Processing Systems}, 2019.

\bibitem{klocek2019hypernetwork}
S.~Klocek, {\L}.~Maziarka, M.~Wo{\l}czyk, J.~Tabor, J.~Nowak, and M.~{\'S}mieja, ``Hypernetwork functional image representation,'' in \emph{International Conference on Artificial Neural Networks}, 2019.

\bibitem{ratzlaff2019hypergan}
N.~Ratzlaff and L.~Fuxin, ``Hypergan: A generative model for diverse, performant neural networks,'' in \emph{International Conference on Machine Learning}, 2019.

\bibitem{lu2023hyperrs}
Y.~Lu, K.~Nakamura, and R.~Ichise, ``{HyperRS}: Hypernetwork-based recommender system for the user cold-start problem,'' \emph{IEEE Access}, 2023.

\bibitem{nemala2023differential}
V.~Nemala, P.~Lai, and N.~Phan, ``Differential privacy in hypernetworks for personalized federated learning,'' in \emph{32nd ACM International Conference on Information and Knowledge Management}, 2023.

\bibitem{kouziokas2020svm}
G.~N. Kouziokas, ``{SVM} kernel based on particle swarm optimized vector and bayesian optimized {SVM} in atmospheric particulate matter forecasting,'' \emph{Applied Soft Computing}, 2020.

\bibitem{ramachandran2017swish}
P.~Ramachandran, B.~Zoph, and Q.~V. Le, ``Searching for activation functions,'' \emph{arXiv preprint arXiv:1710.05941}, 2017.

\bibitem{Miller2020-yc}
C.~Miller, A.~Kathirgamanathan, and P.~et~al., ``The building data genome project 2, energy meter data from the {ASHRAE} great energy predictor {III} competition,'' \emph{Scientific Data}.

\bibitem{fekri2023asynchronous}
M.~Fekri, K.~Grolinger, and S.~Mir, ``Asynchronous adaptive federated learning for distributed load forecasting with smart meter data,'' \emph{International Journal of Electrical Power and Energy Systems}, 2023, accepted.

\bibitem{l2022transformer}
A.~L’Heureux, K.~Grolinger, and M.~A. Capretz, ``Transformer-based model for electrical load forecasting,'' \emph{Energies}, 2022.

\bibitem{torghabeh2023effectiveness}
F.~A. Torghabeh, Y.~Modaresnia, and M.~M. Khalilzadeh, ``Effectiveness of learning rate in dementia severity prediction using {VGG16},'' \emph{Biomedical Engineering: Applications, Basis and Communications}, 2023.

\bibitem{kumar2017weight}
S.~K. Kumar, ``On weight initialization in deep neural networks,'' \emph{arXiv preprint arXiv:1704.08863}, 2017.

\bibitem{qin2022multi}
J.~Qin, Y.~Zhang, S.~Fan, X.~Hu, Y.~Huang, Z.~Lu, and Y.~Liu, ``Multi-task short-term reactive and active load forecasting method based on attention-lstm model,'' \emph{International Journal of Electrical Power \& Energy Systems}, 2022.

\bibitem{bashir2022short}
T.~Bashir, C.~Haoyong, M.~F. Tahir, and Z.~Liqiang, ``Short term electricity load forecasting using hybrid prophet-lstm model optimized by bpnn,'' \emph{Energy reports}, 2022.

\bibitem{klyuev2022methods}
R.~V. Klyuev, I.~D. Morgoev, A.~D. Morgoeva, O.~A. Gavrina, N.~V. Martyushev, E.~A. Efremenkov, and Q.~Mengxu, ``Methods of forecasting electric energy consumption: A literature review,'' \emph{Energies}, 2022.

\end{thebibliography}

\section{Biography Section}

\vspace{-5 cm}

\begin{IEEEbiography}[{\includegraphics[width=1in,height=1.25in,clip,keepaspectratio]{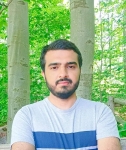}}]{Muhammad Umair Danish}
Muhammad Umair Danish is a PhD student in the Department of Electrical and Computer Engineering at Western University, Canada. He received a BS Degree in Software Engineering from COMSATS University Islamabad. He also received his MS degree in Software Engineering focusing on Machine Learning. His research focuses on Machine Learning, Interpretability of Neural Networks, and Physics-Guided Machine Learning.
\end{IEEEbiography}
\vspace{-5 cm}

\begin{IEEEbiography}[{\includegraphics[width=1in,height=1.25in,clip,keepaspectratio]{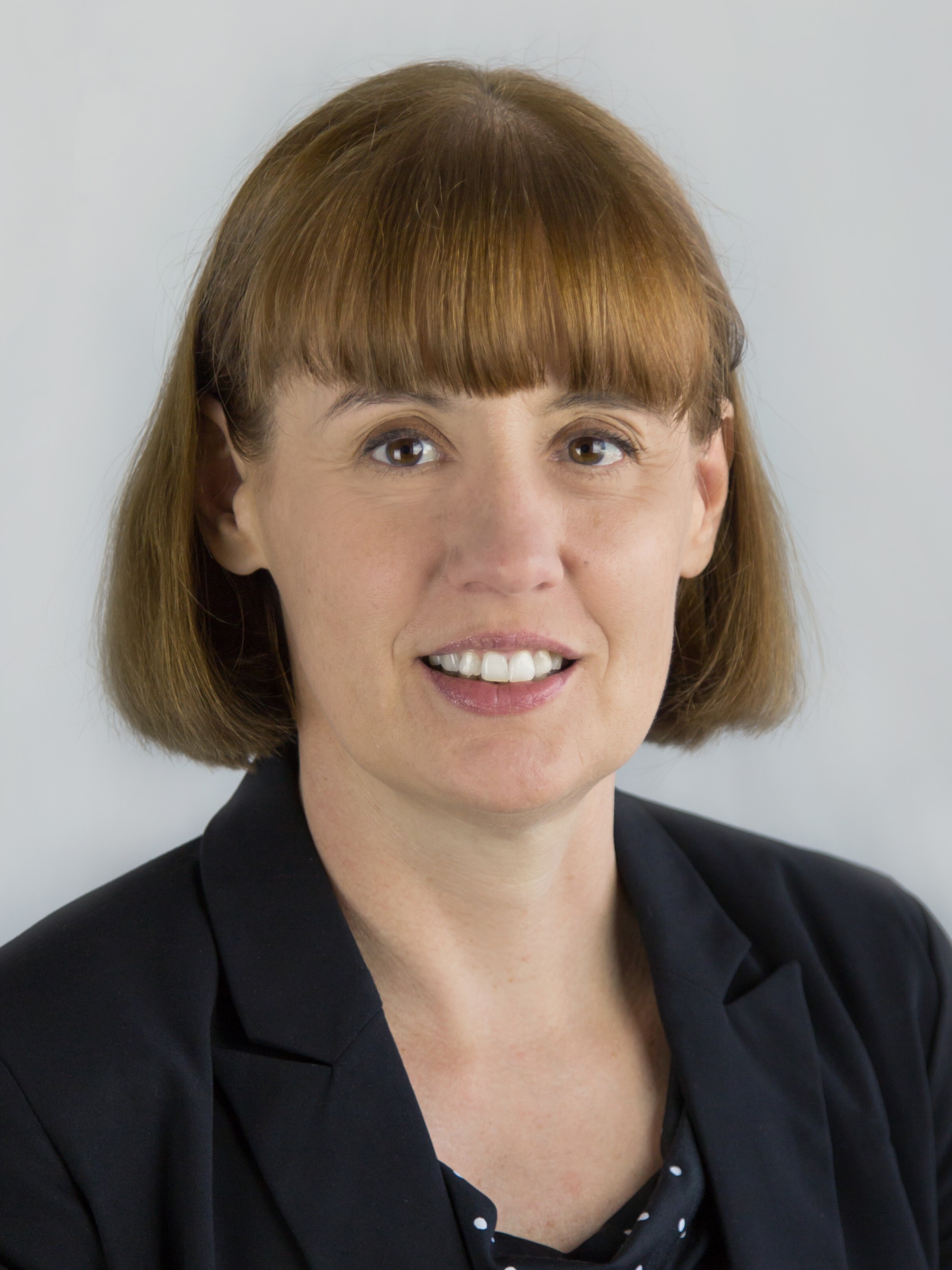}}]{Katarina Grolinger (M'11-SM'24)}
Katarina Grolinger (M’11-SM’24) is an Associate Professor of software engineering in the Department of Electrical and Computer Engineering at Western University, Canada, Canada Research Chair in Engineering Applications of Machine Learning, and a faculty affiliate at Vector institute for AI. Dr. Grolinger received the BSc
and MSc degrees in mechanical engineering from the University of Zagreb, Croatia, and the M.Eng. and PhD degrees in software engineering from Western University, London, Canada.
She has been involved in the software engineering area in academia
and industry, for over 20 years. Her research interests include machine learning, sensor data analytics, data management, and IoT.
\end{IEEEbiography}

\end{document}